\documentclass[runningheads]{llncs}

\usepackage{eccv}

\usepackage{eccvabbrv}

\usepackage{graphicx}
\usepackage{booktabs}
\usepackage[utf8]{inputenc} 
\usepackage[T1]{fontenc}    
\usepackage{url}            
\usepackage{amsfonts}       
\usepackage{nicefrac}       
\usepackage{microtype}      
\usepackage{color}
\usepackage{colortbl}
\usepackage{xcolor}         
\usepackage{multirow}
\usepackage{multicol}
\usepackage{diagbox}
\usepackage{wrapfig}
\usepackage{amsmath,amssymb}
\usepackage{multirow}
\usepackage{makecell}
\usepackage{appendix}
\usepackage{subcaption}
\usepackage{wrapfig}
\usepackage{algorithm}
\usepackage{algorithmic}

\usepackage[accsupp]{axessibility}  

\usepackage{hyperref}

\usepackage{orcidlink}

\begin{document}

\title{Hierarchical Gaussian Mixture Normalizing Flow Modeling for Unified Anomaly Detection} 

\titlerunning{Hierarchical Gaussian Mixture Modeling for Unified Anomaly Detection}

\author{Xincheng Yao\inst{1} \and
Ruoqi Li\inst{1} \and
Zefeng Qian\inst{1} \and Lu Wang\inst{3} \and Chongyang Zhang\inst{1,2\thanks{Corresponding Author.}}}

\authorrunning{Yao et al.}

\institute{School of Electronic Information and Electrical Engineering, Shanghai Jiao Tong University \and
MoE Key Lab of Artificial Intelligence, AI Institute, Shanghai Jiao Tong University
 \and
School of Intelligent Manufacturing, Wuxi Vocational College of Science and Technology\\
\email{\{i-Dover, nilponi, zefeng\_qian, sunny\_zhang\}@sjtu.edu.cn$^1$, wlulu0327@163.com$^3$}}

\maketitle

\begin{abstract}
  Unified anomaly detection (AD) is one of the most valuable challenges for anomaly detection, where one unified model is trained with normal samples from multiple classes with the objective to detect anomalies in these classes. For such a challenging task, popular normalizing flow (NF) based AD methods may fall into a ``homogeneous mapping'' issue, where the NF-based AD models are biased to generate similar latent representations for both normal and abnormal features, and thereby lead to a high missing rate of anomalies. In this paper, we propose a novel \textbf{H}ierarchical \textbf{G}aussian mixture normalizing flow modeling method for accomplishing unified \textbf{A}nomaly \textbf{D}etection, which we call HGAD. Our HGAD consists of two key components: inter-class Gaussian mixture modeling and intra-class mixed class centers learning. Compared to the previous NF-based AD methods, the hierarchical Gaussian mixture modeling approach can bring stronger representation capability to the latent space of normalizing flows. In this way, we can avoid mapping different class distributions into the same single Gaussian prior, thus effectively avoiding or mitigating the ``homogeneous mapping'' issue. We further indicate that the more distinguishable different class centers, the more conducive to avoiding the bias issue. Thus, we further propose a mutual information maximization loss for better structuring the latent feature space. We evaluate our method on four real-world AD benchmarks, where we can significantly improve the previous NF-based AD methods and also outperform the SOTA unified AD methods. The code will be available at \url{https://github.com/xcyao00/HGAD}.

\end{abstract}

\section{Introduction}
\label{sec:introduction}

\begin{figure*}[ht]
    \centering
    \includegraphics[width=1.0\linewidth]{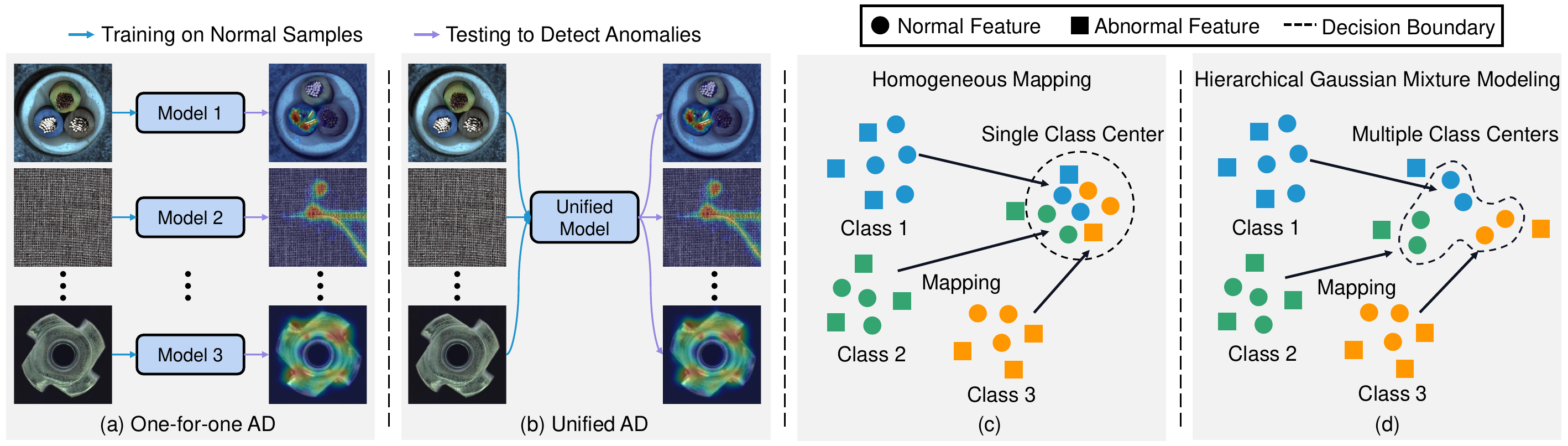}
    \caption{\textbf{Anomaly detection task settings}. We aim to implement one unified AD model (b). (c) Mapping all input features to the same latent class center may induce the ``homogeneous mapping'' issue. (d) We propose a hierarchical Gaussian mixture modeling method for more effectively capturing the complex multi-class distribution.}
    \label{fig:motivation}
\end{figure*}

Anomaly detection has received increasingly wide attentions and applications in different scenarios, such as industrial defect detection \cite{MVTec, PatchCore, UniAD, FastFLOW, RDAD, FOD}, video surveillance \cite{UBnormal, RealWorldAD}, medical lesion detection \cite{Medical1, Medical2}, and road anomaly detection \cite{RoadAD1, RoadAD2}. Considering the highly scarce anomalies and diverse normal classes, most previous AD studies have mainly devoted to unsupervised one-class learning, \emph{i.e.}, learning one specific AD model by only utilizing one-class normal samples and then detecting anomalies in this class\footnote{ Class means the category of the object in the image, for industrial AD, it refers to the industrial product category, e.g., in Fig. \ref{fig:motivation}, cable, carpet, etc.}. However, such a one-for-one paradigm would require more human labor, time, and computation costs when training and testing on many product categories, and also underperform when the one normal class has large intra-class diversity. 
\vspace{-0.3cm}
\vspace{-0.3cm}

In this work, we aim to tackle a more practical task: unified anomaly detection. As shown in Fig. \ref{fig:motivation}b, one unified model is trained with normal samples from multiple classes, and the objective is to detect anomalies for all these classes without any fine-tuning. Nonetheless, solving such a task is quite challenging. Currently, there are two reconstruction-based AD methods for tackling the challenging unified AD task, UniAD \cite{UniAD} and PMAD \cite{PMAD}. But the reconstruction-based methods may fall into the ``identical shortcut reconstruction'' dilemma \cite{UniAD}, where anomalies can also be well reconstructed, resulting in the failure of anomaly detection. UniAD and PMAD attempt to mask the adjacent or suspicious anomalies to avoid identical reconstruction. However, due to the diverse scale and shape of anomalies, the masking mechanism cannot completely avoid the abnormal information leakage during reconstruction, the risk of identical reconstruction is still existing. To this end, we consider designing the unified AD model from the normal data distribution learning perspective. The advantage is that we will no longer face the abnormal information leakage risk in principle, as our method is based on normal data distribution and radically avoids reconstruction. Specifically, we employ normalizing flows (NF) to learn the normal data distribution \cite{CFLOW}.



 However, we find that the NF-based AD methods perform unsatisfactorily when applied to the unified AD task. They usually fall into a ``homogeneous mapping'' issue (see Sec. \ref{sec:revisiting}), where the NF-based AD models are biased to generate large log-likelihoods for both normal and abnormal inputs (see Fig. \ref{fig:analysis}b). We explain this issue as: the NF-based AD methods \cite{CFLOW, FastFLOW} employ the uni-modal Gaussian prior to learn the multi-class distribution (multi-modal). This can be seen as learning a mapping from a heterogeneous space to the latent homogeneous space. To learn the mapping well, the network may be prompted to take a bias to concentrate on the coarse-grained common characteristics (\emph{e.g.}, local pixel correlations) and suppress the fine-grained distinguishable characteristics (\emph{e.g.}, semantic content) among different class features \cite{WhyNF}. Consequently, the network homogeneously maps different class features to the close latent embeddings. Thus, even anomalies can obtain large log-likelihoods and become less distinguishable. 

To address this issue, we first empirically confirm that mapping to multi-modal latent distribution is effective to prevent the model from learning the bias (see Fig. \ref{fig:analysis}c). Accordingly, we propose an inter-class Gaussian mixture modeling approach for NF-based AD networks to more effectively capture the complex multi-class distribution. Second, we argue that the inter-class Gaussian mixture modeling can only ensure the features are drawn to the whole distribution but lacks inter-class repulsion, still resulting in a much weaker discriminative ability for different class features. This may cause different class centers to collapse into the same center. To further increase the inter-class discriminability, we propose a mutual information maximization loss to introduce the class repulsion property to the model for better structuring the latent feature space, where the class centers can be pushed away from each other. Third, we introduce an intra-class mixed class centers learning strategy that can urge the model to learn diverse normal patterns even within one class. Finally, we form a hierarchical Gaussian mixture normalizing flow modeling method for unified anomaly detection, which we call HGAD. In summary, we make the following main contributions:

1. We propose a novel unified AD method: HGAD, where the hierarchical Gaussian mixture modeling approach can bring stronger representation capability to the latent space of normalizing flows for accomplishing the unified AD task. 

2. We specifically propose three key designs: inter-class Gaussian mixture modeling, mutual information maximization loss, and intra-class mixed class centers learning strategy.

3. Under the unified AD task, our method can dramatically improve the unified AD performance of the previous one-for-one NF-based AD methods (\emph{e.g}, CFLOW-AD), boosting the AUROC from 89.0\%/94.0\% to 98.4\%/97.9\%, and also outperform the SOTA unified AD methods (\emph{e.g.}, UniAD).


\section{Related Work}
\label{sec:related_work}

\textbf{Anomaly Detection.} 1) \emph{Reconstruction-based approaches} are the most popular AD methods. These methods rely on the assumption that models trained by normal samples would fail in abnormal image regions. Many previous works attempt to train AutoEncoders \cite{MemoryAE, DRAEM}, Variational AutoEncoders \cite{VAE1} and GANs \cite{AnoGAN, GANomaly} to reconstruct the input images. However, these methods face the ``identical shortcut'' problem \cite{UniAD}. 2) \emph{Embedding-based approaches} recently show better AD performance by using ImageNet pre-trained networks as feature extractors \cite{DeepKNN, SPADE}. PaDiM \cite{PaDiM} extract pre-trained features to model Multivariate Gaussian distribution for normal samples, then utilize Mahalanobis distance to measure the anomaly scores. PatchCore \cite{PatchCore} extends on this line by utilizing locally aggregated features and introducing greedy coreset subsampling to form nominal feature banks. 3) \emph{Knowledge distillation} assumes that the student trained to learn the teacher on normal samples could only regress normal features but fail in abnormal features \cite{STAD}. Recent works mainly focus on feature pyramid \cite{MKD, STPM}, reverse distillation \cite{RDAD}, and asymmetric distillation \cite{AST}. 4) \emph{Unified AD approaches} attempt to train a unified AD model to accomplish anomaly detection for multiple classes. UniAD \cite{UniAD}, PMAD \cite{PMAD} and OmniAL \cite{OmniAL} are three existing methods in this new direction. UniAD is a transformer-based reconstruction model with three improvements, it can perform well under the unified case by addressing the ``identical shortcut'' issue. PMAD is a MAE-based patch-level reconstruction model, which can learn a contextual inference relationship within one image rather than the class-dependent reconstruction mode. OmniAL is a unified CNN framework with anomaly synthesis, reconstruction and localization improvements. 


 \textbf{GMM in Anomaly Detection.} Previous methods, DAGMM \cite{DAGMM} and PEDENet \cite{PEDENet}, also utilize GMM for anomaly detection. However, our method is significantly different from these methods as follows: 1) Different Task. DAGMM is used for non-image data, and PEDENet is used for one-for-one image anomaly detection. Although they both use GMM, it's nontrivial to use them for unified anomaly detection. However, our method can achieve superiority in the unified AD task. 2) Different Approach. DAGMM and PEDENet both predict membership and then use the predicted membership and GMM modeling formulas to directly calculate the mixture component weights and the mean and variance of each mixture component. Our method is based on the learnable class centers and class weights. The way of ours is more beneficial for the hierarchical Gaussian mixture modeling while the way in DAGMM and PEDENet is difficult to achieve this.

 \textbf{Normalizing Flows in Anomaly Detection.} In anomaly detection, normalizing flows are employed to learn the normal data distribution \cite{DifferNet, CFLOW, FastFLOW, BGAD, FOOD}, which maximize the log-likelihoods of normal samples during training. Rudolph \emph{et al.} \cite{DifferNet} first employ NFs for anomaly detection by estimating the distribution of pre-trained features. In CFLOW-AD \cite{CFLOW}, the authors further construct NFs on multi-scale feature maps to achieve anomaly localization. Recently, fully convolutional normalizing flows \cite{CSFLOW, FastFLOW} have been proposed to improve the accuracy and efficiency of anomaly detection. In BGAD \cite{BGAD}, the authors propose a NF-based AD model to tackle the supervised AD task. In this paper, we mainly propose a novel NF-based AD model (HGAD) with three improvements to achieve much better unified AD performance.

\section{Method}
\label{sec:method}

\subsection{Preliminary of Normalizing Flow Based Anomaly Detection}
\label{sec:nf_base}

In subsequent sections, upper case letters denote random variables (RVs) (\emph{e.g.}, $X$) and lower case letters denote their instances (\emph{e.g.}, $x$). The probability density function of a RV is written as $p(X)$, and the probability value for one instance as $p_X(x)$. The normalizing flow models \cite{realNVP, Glow} can fit an arbitrary distribution $p(X)$ by a tractable latent base distribution with $p(Z)$ density and a bijective invertible mapping $\varphi: X \in \mathbb{R}^d \rightarrow Z \in \mathbb{R}^d$. Then, according to the change of variable formula \cite{VariationRule}, the log-likelihood of any $x \in X$ can be estimated as:
\begin{equation}
\label{eq:formula1}
    {\rm log}p_\theta(x) = {\rm log}p_Z(\varphi_\theta(x)) + {\rm log}|{\rm det}J|
\end{equation}
where $\theta$ means the learnable model parameters, and we use $p_\theta(x)$ to denote the estimated probability value of feature $x$ by the model $\varphi_\theta$. The $J = \bigtriangledown_x\varphi_\theta(x)$ is the Jacobian matrix of the bijective transformation ($z = \varphi_\theta(x)$ and $x = \varphi_\theta^{-1}(z)$). The model parameters $\theta$ can be optimized by maximizing the log-likelihoods across the training distribution $p(X)$. The loss function is defined as:
\begin{equation}
\label{eq:formula2}
    \mathcal{L}_m = \mathbb{E}_{x \sim p(X)}[-{\rm log}p_\theta(x)]
\end{equation}

In anomaly detection, the latent variables $Z$ for normal features are usually assumed to obey $\mathcal{N}(0, \mathbb{I})$ for simplicity \cite{DifferNet}. By replacing $p_Z(z) = (2\pi)^{-\frac{d}{2}}{\rm e}^{-\frac{1}{2}z^Tz}, z = \varphi_\theta(x)$ in Eq. \ref{eq:formula1}, the loss function in Eq. \ref{eq:formula2} can be written as:
\begin{equation}
\label{eq:formula3}
    \mathcal{L}_m = \mathbb{E}_{x \sim p(X)}\Big[\frac{d}{2}{\rm log}(2\pi) + \frac{1}{2}\varphi_\theta(x)^T\varphi_\theta(x) - {\rm log}|{\rm det}J|\Big]
\end{equation}

After training, the log-likelihoods of the input features can be exactly estimated by the trained normalizing flow models as ${\rm log}p_\theta(x) = -\frac{d}{2}{\rm log}(2\pi) - \frac{1}{2}\varphi_\theta(x)^T\varphi_\theta(x) + {\rm log}|{\rm det}J|$. Next, we can convert log-likelihoods to likelihoods via exponential function: $p_\theta(x) = {\rm e}^{{\rm log}p_\theta(x)}$. As we maximize log-likelihoods for normal features in Eq. \ref{eq:formula2}, the estimated likelihoods $p_\theta(x)$ can directly represent the normality (a larger value means more normal). Thus, we can convert likelihoods to anomaly scores by $s(x) = 1 - p_\theta(x)$ \cite{BGAD}.

\subsection{Revisiting Normalizing Flow Based Anomaly Detection Methods}
\label{sec:revisiting}

\begin{figure*}[ht]
    \centering
    \includegraphics[width=1.0\linewidth]{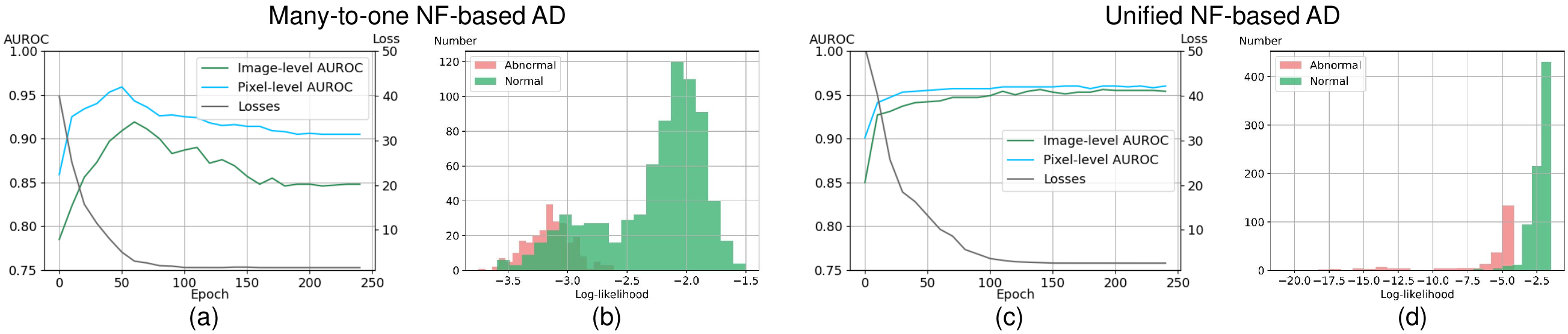}
    \caption{\textbf{Comparison between many-to-one and our unified NF-based AD methods on MVTecAD}. (a) and (c) show the training losses and the testing anomaly detection and localization AUROCs. (b) shows that the many-to-one NF-based AD model (\emph{i.e.}, training the one-for-one AD method, FastFlow, on multiple classes simultaneously) may have an obvious normal-abnormal overlap under the unified case, while ours (d) can bring better normal-abnormal distinguishability.}
    \label{fig:analysis}
\end{figure*}

Under the unified AD task, we follow the NF-based anomaly detection paradigm \cite{DifferNet, CFLOW} and reproduce the FastFlow \cite{FastFLOW} to estimate log-likelihoods of the features extracted by a pre-trained backbone. We then convert the estimated log-likelihoods to anomaly scores and evaluate the AUROC metric every $10$ epochs. As shown in Fig. \ref{fig:analysis}a, after a period of training, the performance of the model drops severely while the losses continue going extremely small. Accordingly, the overall log-likelihoods become much large. We attribute this phenomenon to the ``homogeneous mapping'' issue, where the normalizing flows may map all inputs to much close latent variables and then present large log-likelihoods for both normal and abnormal features, thus failing to detect anomalies. This speculation is empirically verified by the visualization results in Fig. \ref{fig:analysis}b (more results in App. Fig. \ref{fig:logp_results_sup}), where the normal and abnormal log-likelihoods are highly overlapped. As we explained in Sec. \ref{sec:introduction}, the phenomenon may come from that the model excessively suppresses the fine-grained distinguishable characteristics between normal and abnormal features. However, as shown in Fig. \ref{fig:analysis}c and \ref{fig:analysis}d, our unified NF-based AD method can more effectively avoid highly overlapped normal and abnormal log-likelihoods, indicating a slighter log-likelihoods bias problem. This encourages us to analyze as follows.

\vspace{-0.3cm}
\vspace{-0.3cm}

Below, we denote normal features as $x_n \in \mathbb{R}^d$ and abnormal features as $x_a \in \mathbb{R}^d$, where $d$ is the channel dimension. We provide a rough analysis using a simple one coupling layer normalizing flow model. When training, the forward affine coupling \cite{realNVP} can be calculated as:
\begin{equation}
\label{eq:coupling_transformation}
    x_1, x_2 = {\rm split}(x_n) \; ; \; z_1 = x_1, z_2 = x_2 \odot {\rm exp}(s(x_1)) + t(x_1) \; ; \; z = {\rm cat}(z_1, z_2)
\end{equation}
where ${\rm split}$ and ${\rm cat}$ mean split and concatenate the feature maps along the channel dimension, $s(x_1)$ and $t(x_1)$ are transformation coefficients predicted by a learnable neural network \cite{realNVP}. With the maximum likelihood loss in Eq. \ref{eq:formula3} pushing all $z$ to fit $\mathcal{N}(0, \mathbb{I})$, the model has no need to distinguish different class features. Thus, it is more likely to take a bias to predict all $s(\cdot)$ to be very small negative numbers ($\rightarrow -\infty$) and $t(\cdot)$ close to zero. The impact is that the model could also fit $x_a$ to $\mathcal{N}(0, \mathbb{I})$ well with the bias, failing to detect anomalies. However, if we map different class features to different class centers, the model is harder to simply take a bias solution. Instead, $s(\cdot)$ and $t(\cdot)$ must be highly related to input features. Considering that $s(\cdot)$ and $t(\cdot)$ in the trained model are relevant to normal features, the model thus could not fit $x_a$ well. We think that the above rough analysis can also be applied to multiple layers. Because the output of one coupling layer will tend to 0 when 
 $s(\cdot)$ and $t(\cdot)$ of the layer are biased. From Eq. \ref{eq:coupling_transformation}, we can see that when the output of one coupling layer is close to $0$, the output of the next layer will also tend to $0$. Therefore, the output after multiple layers will tend to $0$, the network is still biased.


\subsection{Hierarchical Gaussian Mixture Normalizing Flow Modeling}

\begin{figure*}[ht]
    \centering
    \includegraphics[width=1.0\linewidth]{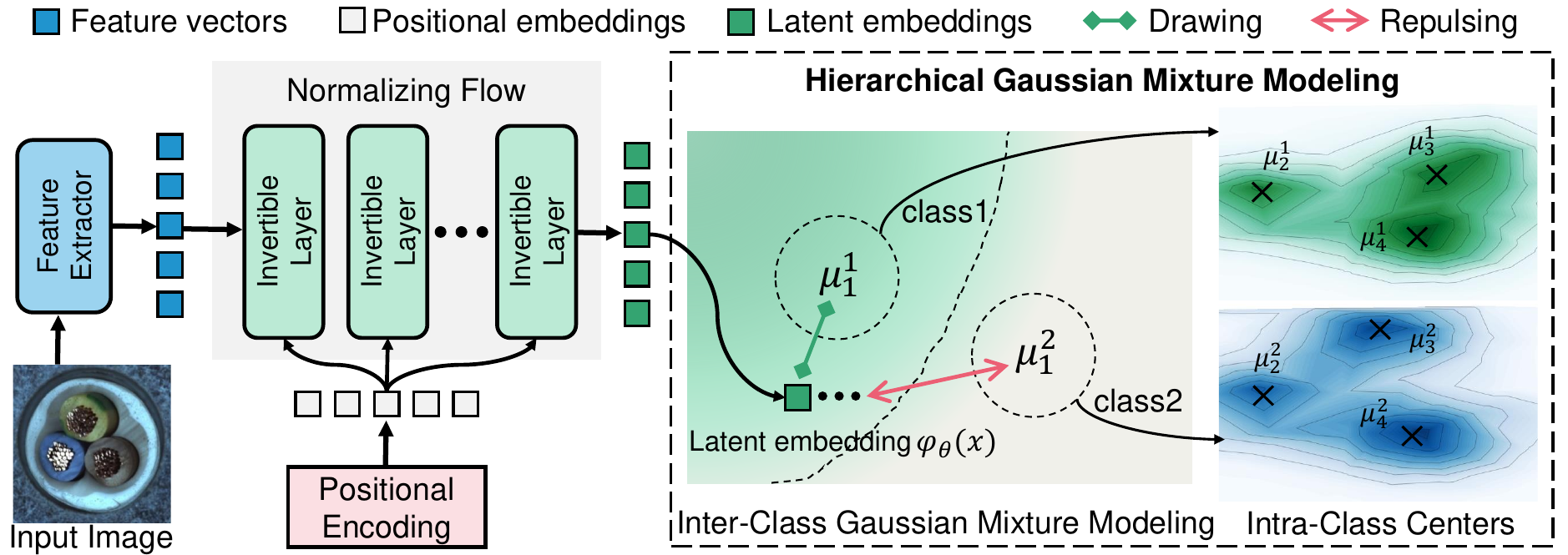}
    \caption{\textbf{Model overview}. The extracted feature vectors are sent into the normalizing flow model for transforming into latent embeddings. Following \cite{CFLOW}, we add positional embeddings (\emph{i.e.}, sinusoidal position encoding) to each invertible layer as they are effective for NF-based AD methods. Then, we employ our hierarchical Gaussian mixture modeling approach to fit the latent embeddings, which can assist the model against learning the ``homogeneous mapping''.}
    \label{fig:framework}
\end{figure*}

\textbf{Overview.} As shown in Fig. \ref{fig:framework}, our HGAD is composed of a feature extractor, a normalizing flow model (details in App. \ref{sec:implementation_sup}), and the hierarchical Gaussian mixture modeling. First, the features extracted by a fixed pre-trained backbone are sent into the normalizing flow model to transform into the latent embeddings. Then, we employ our hierarchical Gaussian mixture modeling method to fit the latent embeddings during training. In Fig. \ref{fig:framework}, we only show features from one level, but we indeed extract multi-level feature maps and construct the NF model at each level. The pseudo-code of our HGAD is provided in Alg. \ref{algorithm1}. 

\vspace{-0.3cm}
\vspace{-0.3cm}


\begin{algorithm}
	\renewcommand{\algorithmicrequire}{\textbf{Input:}}
	\renewcommand{\algorithmicensure}{\textbf{Output:}}
	\caption{HGAD: \textbf{H}ierarchical \textbf{G}aussian mixture modeling for unified \textbf{AD}}
	\label{algorithm1}
	\begin{algorithmic}[1]
     \REQUIRE Input image $I \in \mathbb{R}^{H\times W \times 3}$, Class $y \in \{1,\dots,Y\}$ 
		\STATE Initialization: Class centers $\mu^k_y \leftarrow y + \boldsymbol{r}, \boldsymbol{r} \in \mathbb{R}^{d_k} \sim \mathcal{N}(0,\mathbb{I})$, Learnable vector $\psi^k \leftarrow \boldsymbol{0} \in \mathbb{R}^{Y}$, $\psi^k_y \leftarrow \boldsymbol{0} \in \mathbb{R}^{M}$; $y \in \{1,\dots,Y\}, k \in \{1,\dots,K\}$
		\STATE Feature Extraction: $K$-level feature maps, denoted as $X^k, k \in \{1,\dots,K\}$
      \FOR {each level $k \in \{1,\dots,K\}$} 
      \STATE Total loss $\mathcal{L}_{all} \leftarrow 0$
      \FOR {each feature $x \in X^k$} 
      \STATE Obtain the latent representation: $\varphi_\theta(x) \in \mathbb{R}^{d_k}$
      \STATE Calculate $c_y = {\rm logsoftmax}_y(\psi^k)$ and $c_i^y = {\rm logsoftmax}_i(\psi^k_y), i \in \{1,\dots,M\}$
      \STATE Calculate \textbf{inter-class loss}: $\mathcal{L}_g$ based on Eq. \ref{eq:loss_x}
      \STATE Calculate \textbf{mutual information maximization loss}: $\mathcal{L}_{mi}$ based on Eq. \ref{eq:loss_y2}
      \STATE Calculate \textbf{entropy loss}: $\mathcal{L}_e$ based on Eq. \ref{eq:loss_e}
      \STATE Calculate \textbf{intra-class loss}: $\mathcal{L}_{in}$ based on Eq. \ref{eq:loss_x_intra}
      \STATE Calculate the overall loss: $\mathcal{L} = \lambda_1 \mathcal{L}_g + \lambda_2 \mathcal{L}_{mi} + \mathcal{L}_e + \mathcal{L}_{in}$
      \STATE Update $\mathcal{L}_{all} \leftarrow \mathcal{L}_{all} + \mathcal{L}$
      \ENDFOR
      \STATE Obtain mean loss $\mathcal{L}^k_{mean} = \mathcal{L}_{all} / N$, $N$ is the total number of features  
      \ENDFOR
		\ENSURE  Mean loss $\mathcal{L}_{mean} = \frac{1}{K}\sum_{k=1}^K\mathcal{L}^k_{mean}$
	\end{algorithmic}  
\end{algorithm}


\textbf{Inter-Class Gaussian Mixture Modeling.} As discussed in Sec. \ref{sec:revisiting}, mapping different class features to different class centers can suffer a slighter log-likelihoods bias problem. To this end, to further better fit the complex multi-class normal distribution in the latent space, we propose the inter-class Gaussian mixture modeling approach. Specifically, a Gaussian mixture model with class-dependent means $\mu_y$ and covariance matrices $\Sigma_y$ ($y$ is used to distinguish different classes), is used as the prior distribution for the latent variables $Z$:
\begin{equation}
\label{eq:gmm}
    p_{Z|Y}(z|y) = \mathcal{N}(z; \mu_y, \Sigma_y)\quad {\rm and}\quad  p_Z(z) = \sum\nolimits_y p(y)\mathcal{N}(z; \mu_y, \Sigma_y)
\end{equation}

For simplicity, we also use unit matrix $\mathbb{I}$ to replace all the class-dependent covariance matrices $\Sigma_y$. To urge the network to adaptively learn the class weights, we parameterize the class weights $p(Y)$ through a learnable vector $\psi$, with $p(y) = {\rm softmax}_y(\psi)$, where the subscript $y$ of the softmax operator denotes the class index of the calculated softmax value. The use of the softmax can ensure that $p(y)$ stays positive and sums to one. The $\psi$ can be initialized to $0$. With the parameterized $p(Y)$, we can derive the loss function for the inter-class Gaussian mixture modeling as follows (the detailed derivation is in App. \ref{sec:app_loss_derivation}):
\begin{equation}
\label{eq:loss_x}
    \mathcal{L}_g = \mathbb{E}_{x \sim p(X)}\bigg[-\mathop{{\rm logsumexp}}\limits_y\bigg(-\frac{||\varphi_\theta(x)-\mu_y||^2_2}{2}+c_y\bigg)-{\rm log}|{\rm det}J| + \frac{d}{2}{\rm log}(2\pi)\bigg]
\end{equation}
where $c_y$ denotes logarithmic class weights and is defined as $c_y := {\rm log}p(y) = {\rm logsoftmax}_y(\psi)$, and the subscript $y$ of the ${\rm logsumexp}$ operator denotes summing the ${\rm exp}$ values of all classes.

\textbf{Mutual Information Maximization.} Next, we further argue that the inter-class Gaussian mixture modeling can only ensure the latent features are drawn together to the whole distribution (parameterized by $\{\mu_y, \psi_y\}_{y=1}^Y$), where the $\{p(y)\}_{y=1}^Y$ can control the contribution of different class centers to the log-likelihood estimation value ${\rm log}p_\theta(x)$. This means that the loss function in Eq. \ref{eq:loss_x} only has the drawing characteristic to make the latent features fit the multi-modal distribution, but without the repulsion property for separating among different classes, still resulting in a much weaker discriminative ability for different class features. As the class centers are randomly initialized, this may cause different class centers to collapse into the same center. To address this, we consider that the latent feature $z$ from class $y$ should be drawn close to its corresponding class center $\mu_y$ as much as possible while far away from the other class centers. From the information theory perspective, this means that the mutual information $I(Y, Z)$ should be large enough. So, we propose a mutual information maximization loss to introduce the class repulsion property for increasing the class discrimination ability. The loss function is defined as follows (the derivation is in App. \ref{sec:app_loss_derivation}):
\begin{equation}
\label{eq:loss_y1}
    \mathcal{L}_{mi} = -\mathbb{E}_{y \sim p(Y)}[-{\rm log}p(y)] - \mathbb{E}_{(x,y) \sim p(X,Y)}\bigg[{\rm log}\frac{p(y)p(\varphi_\theta(x)|y)}{\sum_{y^\prime}p(y^\prime)p(\varphi_\theta(x)|y^\prime)}\bigg]
\end{equation}

 By replacing $p(\varphi_\theta(x)|y)$ with $\mathcal{N}(\varphi_\theta(x); \mu_y, \mathbb{I})$ in Eq. \ref{eq:loss_y1}, we can derive the following practical loss format (the detailed derivation is in App. \ref{sec:app_loss_derivation}):
\begin{equation}
\label{eq:loss_y2}
    \mathcal{L}_{mi} = -\mathbb{E}_{(x,y) \sim p(X,Y)}\bigg[\mathop{{\rm logsoftmax}}\limits_{y}\bigg(-\frac{||\varphi_\theta(x)-\mu_{y^\prime}||^2_2}{2} + c_{y^\prime}\bigg) - c_y\bigg]
\end{equation}
where $\mu_{y\prime}$ means all the other class centers except for $\mu_y$, $c_{y\prime}$ means all the other logarithmic class weights except for $c_y$, and the subscript $y$ of the ${\rm logsoftmax}$ denotes calculating logsoftmax value for class $y$. Note that we also use this representation way for softmax calculation in the following sections (\emph{e.g.}, Eq. \ref{eq:loss_e}).

In addition to the mutual information maximization loss, we propose that we can also introduce the class repulsion property by minimizing the inter-class entropy. We use the $-||\varphi_\theta(x)-\mu_y||^2_2/2$ as the class logits for class $y$, and then define the entropy loss as follows (a standard entropy formula):
\begin{equation}
\label{eq:loss_e}
    \mathcal{L}_e = \mathbb{E}_{x \sim p(X)}\bigg[\sum_y -\mathop{{\rm softmax}}\limits_y(-||\varphi_\theta(x)-\mu_{y^\prime}||^2_2/2) \cdot \mathop{{\rm logsoftmax}}\limits_y(-||\varphi_\theta(x)-\mu_{y^\prime}||^2_2/2)\bigg]
\end{equation}

\textbf{Learning Intra-Class Mixed Class Centers.} In real-world scenarios, even one object class may contain diverse normal patterns. Thus, to better model intra-class distribution, we further extend the Gaussian prior $p(Z|y) = \mathcal{N}(\mu_y, \Sigma_y)$ to mixture Gaussian prior $p(Z|y) = \sum\nolimits_{i=1}^Mp_i(y)\mathcal{N}(\mu_i^y, \Sigma^y_i)$, where $M$ is the number of intra-class latent centers. We can directly replace the $p(Z|y)$ in Eq. \ref{eq:gmm} and derive the corresponding loss function $\mathcal{L}_g$ in Eq. \ref{eq:loss_x} (see App. Eq. \ref{eq:loss_gmm_intra}). However, the initial latent features $Z$ usually have large distances with the intra-class centers $\{\mu_i^y\}_{i=1}^M$, this will cause the $p(z|y), z \in Z$ close to $0$. After calculating the logarithm function, it is easy to cause the loss to be numerically ill-defined (NaN), making it fail to be optimized. To this end, we propose to decouple the inter-class Gaussian mixture modeling and the intra-class latent centers learning. This decoupling strategy is more conducive to learn class centers as we form a coarse-to-fine optimization process. Specifically, for each class $y$, we learn a main class center $\mu_1^y$ and the delta vectors $\{\Delta\mu^y_i\}_{i=1}^M$ ($\Delta\mu^y_1$ is fixed to $0$), which mean the offset values from the main center and are used to represent the other intra-class centers: $\mu^y_i=\{\mu^y_1+\Delta\mu^y_i\}_{i=1}^M$. Then, we can directly employ the Eq. \ref{eq:loss_x} to optimize the main center $\mu_1^y$. When learning the other intra-class centers, we detach the main center $\mu_1^y$ from the gradient graph and only optimize the delta vectors by the following loss function:
\begin{equation}
\label{eq:loss_x_intra}
    \mathcal{L}_{in} = \mathbb{E}_{(x,y) \sim p(X,Y)}\bigg[-\mathop{{\rm logsumexp}}\limits_i\bigg(-\frac{||\varphi_\theta(x)-(SG[\mu^y_1]+\Delta\mu^y_i)||^2_2}{2}+c^y_i\bigg)-{\rm log}|{\rm det}J|\bigg]
\end{equation}
where $SG[\cdot]$ means to stop gradient backpropagation, $c_i^y$ denotes logarithmic intra-class center weights and is defined as $c_i^y := {\rm log}p_i(y) = {\rm logsoftmax}_i(\psi_y)$. Note that $\psi_y \in \mathbb{R}^{M}$ is specific to class $y$, $\psi \in \mathbb{R}^{Y}$ is for the whole distribution.

\textbf{Overall Loss Function.} The overall training loss function is the combination of the Eq. \ref{eq:loss_x}, Eq. \ref{eq:loss_y2}, Eq. \ref{eq:loss_e} and Eq. \ref{eq:loss_x_intra}, as follows:
\begin{equation}
\label{eq:loss_overall}
    \mathcal{L} = \lambda_1 \mathcal{L}_g + \lambda_2 \mathcal{L}_{mi} + \mathcal{L}_e + \mathcal{L}_{in}
\end{equation}
where the $\lambda_1$ and $\lambda_2$ are used to trade off the loss items and are set to $1$ and $100$ by default. The $\mathcal{L}_e$ and $\mathcal{L}_{in}$ are used as auxiliary losses, so we don't ablate weighting factors for them as it will result in too many combinational experiments. In App. \ref{sec:motivation}, we further explain the necessities and motivations that connect each other of the individual components in our method.


\subsection{Anomaly Scoring}
\label{sec:anomaly_scoring}

 As shown in Alg. \ref{algorithm1}, we actually extract K-level feature maps and construct the NF model at each level. For each test input feature $x^k$ from level-$k$, $k \in K$, we can calculate its intra-class log-likelihood ${\rm log}p_\theta(x^k) = \mathop{{\rm logsumexp}}_i(-||\varphi_\theta(x^k)-\mu_i^y||^2_2/2+c_i^y)+{\rm log}|{\rm det}J| - d/2{\rm log}(2\pi)$ and inter-class negative entropy $nh(x^k) = \sum_y \mathop{{\rm softmax}}_y(-||\varphi_\theta(x^k)-\mu_1^{y^\prime}||^2_2/2) \cdot \mathop{{\rm logsoftmax}}_y(-||\varphi_\theta(x^k)-\mu_1^{y^\prime}||^2_2/2)$. Note that $y$ in $c_i^y$ denotes which class $x^k$ belongs to, not whether $x^k$ is normal or abnormal. Next, we convert the log-likelihood to likelihood $p_\theta(x^k) = {\rm e}^{{\rm log}p_\theta(x^k)}$. Then, we upsample all $p_\theta(x^k)$ in the level-$k$ to the input image resolution $(H \times W)$ using bilinear interpolation $P_k = b(p_\theta(x^k)) \in \mathbb{R}^{H \times W}$. Finally, we calculate anomaly score map $S_l$ by aggregating all upsampled likelihoods as $S_l = 1 - \sum_{k=1}^K P_k$ (see Sec. \ref{sec:nf_base}). For the inter-class negative entropy, we also follow the above steps to convert to anomaly score map $S_e$. Then the final anomaly score map is obtained by combining the two maps $S = S_l \odot S_e$, where the $\odot$ is the element-wise multiplication. In this way, even if anomalies fall into the inter-class Gaussian mixture distribution, they are usually in the low-density regions among the inter-class class centers. So, we can still ensure that anomalies are out-of-distribution through intra-class log-likelihoods (please see App. \ref{sec:guaratee_way} for more discussions).


\subsection{Further Discussions}

\textbf{Explicitly Distinguishing Classes.} Our method utilizes the class $y$ of each image, but it doesn't introduce any extra data collection cost compared to one-for-one AD models. The existing AD datasets are collected for one-for-one anomaly detection (\emph{i.e.}, we need to train a model for each class). Thus, the existing AD datasets
need to be separated according to classes, with each class as a subdataset. Therefore, one-for-one AD methods also need to distinguish classes, as they require normal samples from the same class to train. Our method actually has the same supervision as these methods. The difference is that we explicitly distinguish classes but they don’t explicitly distinguish classes. So, our method still follows the same data organization format as the one-for-one AD models. But the advantage of our unified AD method is that we can train one model for all classes, greatly reducing the resource costs of training and deploying. Moreover, our method only requires separating different classes and does not require pixel-level annotations. For real-world industrial applications, this doesn't incur extra data collection costs, as we usually consciously collect data according to different classes. In App. \ref{sec:sup_info}, we summarize the training samples and the supervision information required by our method and other methods in detail. PMAD \cite{PMAD} uses class information, and both UniAD \cite{UniAD} and OmniAL \cite{OmniAL} use extra information to simulate anomalies. Although we also use class information, our method can achieve superior results than these methods. Thus, we think that whether utilizing class information should not be a strict limitation to the unified AD modeling especially when it already exists.




In App. \ref{sec:discussion_with_identical_shortcut}, we further provide more discussions with the ``identical shortcut'' issue and point out that the ``homogeneous mapping'' is not intrinsically equal to the ``identical shortcut'' issue. Compared to UniAD, our method is an effective exploration for unified anomaly detection in the direction of normalizing flow based anomaly detection. In App. \ref{sec:discussion_sup}, we also further discuss the limitations, applications, effective guarantee, and complexity of our method. We also provide an information-theoretic view in App. \ref{sec:information_view}.

\section{Experiment}
\label{sec:experiment}

\subsection{Datasets and Metrics}

\textbf{Datasets.} We extensively evaluate our approach on four real-world industrial AD datasets: MVTecAD \cite{MVTec}, BTAD \cite{BTAD}, MVTec3D-RGB \cite{MVTec3D}, and VisA \cite{VisA}. The detailed introduction to these datasets is provided in App. \ref{sec:dataset_sup}. To more sufficiently evaluate the unified AD performance of different AD models, we combine these datasets to form a 40-class dataset, which we call Union dataset.

\textbf{Metrics.} Following prior works \cite{MVTec, STAD, DRAEM}, the standard metric in anomaly detection, AUROC, is used to evaluate the performance of AD methods. 

\subsection{Main Results}

\textbf{Setup.} For each dataset, we train one unified model with images from different classes. We use Efficient-b6 \cite{Efficientnet} as the feature extractor. The parameters of the feature extractor are frozen during training. The layer numbers of the NF models are all 12. The number of inter-class centers is always equal to the number of classes in the dataset. The number of intra-class centers is set as $10$ for all datasets (see ablation study in Sec. \ref{sec:ablation_studies}). We use the Adam \cite{Adam} optimizer with weight decay $1{\rm e}^{-4}$ to train the model. The total training epochs are set as 100 and the batch size is 8 by default. The learning rate is $2{\rm e}^{-4}$ initially, and dropped by $0.1$ after $[48,57,88]$ epochs. The evaluation is run with 3 random seeds. 

\textbf{Baselines.} We compare our approach with one-for-one AD baselines including: PaDiM \cite{PaDiM}, MKD \cite{MKD}, and DRAEM \cite{DRAEM}, and SOTA unified AD methods: PMAD \cite{PMAD}, UniAD \cite{UniAD}, and OmniAL \cite{OmniAL}. We also compare with the SOTA one-for-one NF-based AD methods: CFLOW \cite{CFLOW} and FastFlow \cite{FastFLOW}. Under the unified case, the results of the one-for-one AD baselines and the NF-based AD methods are run with the publicly available implementations.

\begin{table*}
\caption{\textbf{Anomaly detection and localization results on MVTecAD}. All methods are evaluated under the unified case. $\cdot/\cdot$ means the image-level and pixel-level AUROCs.}
\label{tab:detailed_MVTecAD}
\resizebox{1.0\linewidth}{!}{
\begin{tabular}{c||ccc|ccc|ccc}
\toprule
\multirow{2}*{\textbf{Category}} & \multicolumn{3}{c|}{\textbf{Baseline Methods}} & \multicolumn{3}{c|}{\textbf{Unified Methods}} & \multicolumn{3}{c}{\textbf{Normalizing Flow Based Methods}}\\
  & PaDiM  & MKD  & DRAEM  & PMAD & UniAD & OmniAL  & FastFlow  & CFLOW  & HGAD (Ours)\\
\midrule
\midrule
 Carpet & 93.8/97.6 & 69.8/95.5 & 98.0/98.6 & 99.0/97.9 & 99.8/98.5 & 98.7/\textbf{99.4} & 91.6/96.7 & 98.8/97.5 & \textbf{100}$\pm$0.00/\textbf{99.4}$\pm$0.05\\
  Grid  & 73.9/71.0 & 83.8/82.3 & 99.3/98.7 & 96.2/95.6 & 98.2/96.5 & \textbf{99.9}/\textbf{99.4} & 85.7/96.8 & 95.9/94.1 & 99.6$\pm$0.09/99.1$\pm$0.08\\
  Leather & 99.9/84.8 & 93.6/96.7 & 98.7/97.3 & \textbf{100}/99.2 & \textbf{100}/98.8 & 99.0/99.3 & 93.7/98.2 & \textbf{100}/98.1 & \textbf{100}$\pm$0.00/\textbf{99.6}$\pm$0.00\\
  Tile & 93.3/80.5 & 89.5/85.3 & 99.8/98.0 & 99.8/94.5 & 99.3/91.8 & 99.6/\textbf{99.0}  & 99.2/95.8 & 97.9/92.2 & \textbf{100}$\pm$0.00/96.1$\pm$0.09\\
  Wood  & 98.4/89.1 & 93.4/80.5 & \textbf{99.8}/96.0 & 99.6/89.0 & 98.6/93.2 & 93.2/\textbf{97.4} & 98.0/92.0 & 99.0/92.7 & 99.5$\pm$0.08/95.9$\pm$0.09\\
\midrule
  Bottle & 97.9/96.1 & 98.7/91.8 & 97.5/87.6 & 99.8/98.4 & 99.7/98.1 & \textbf{100}/\textbf{99.2} & \textbf{100}/94.0 & 98.7/96.4 & \textbf{100}$\pm$0.00/98.6$\pm$0.08\\
 Cable & 70.9/81.0 & 78.2/89.3 & 57.8/71.3 & 93.5/95.4 & 95.2/\textbf{97.3} & \textbf{98.2}/\textbf{97.3} & 90.9/95.2 & 80.4/92.9 & 97.3$\pm$0.26/95.2$\pm$0.49\\
 Capsule & 73.4/96.9 & 68.3/88.3 & 65.3/50.5 & 80.5/97.0 & 86.9/98.5 & 95.2/96.9 & 90.5/98.6 & 75.5/97.7 & \textbf{99.0}$\pm$0.40/\textbf{99.2}$\pm$0.05\\
 Hazelnut & 85.5/96.3 & 97.1/91.2 & 93.7/96.9 & 99.6/97.4 & 99.8/98.1 & 95.6/98.4 & 98.9/96.6 & 97.1/95.7 & \textbf{99.9}$\pm$0.08/\textbf{98.8}$\pm$0.05\\
 Metal nut & 88.0/84.8 & 64.9/64.2 & 72.8/62.2 & 98.0/91.7 & 99.2/94.8 & 99.2/\textbf{99.1} & 96.5/97.2 & 87.8/84.4 & \textbf{100}$\pm$0.00/97.8$\pm$0.29\\
 Pill & 68.8/87.7 & 79.7/69.7 & 82.2/94.4 & 89.4/93.4 & 93.7/95.0 & \textbf{97.2}/\textbf{98.9} & 90.4/96.1 & 88.0/90.7 & 96.3$\pm$0.73/98.8$\pm$0.05\\
 Screw & 56.9/94.1 & 75.6/92.1 & 92.0/95.5 & 73.3/96.6 & 87.5/98.3 & 88.0/98.0 & 76.8/95.9 & 59.5/93.9 & \textbf{95.5}$\pm$0.16/\textbf{99.3}$\pm$0.12\\
 Toothbrush & 95.3/95.6 & 75.3/88.9 & 90.6/97.7 & 95.8/98.2 & 94.2/98.4 & \textbf{100}/\textbf{99.4} & 86.1/97.1 & 78.0/95.7 & 91.2$\pm$0.37/99.1$\pm$0.05\\
 Transistor & 86.6/92.3 & 73.4/71.7 & 74.8/64.5 & 97.2/93.3 & \textbf{99.8}/\textbf{97.9} & 93.8/93.3 & 85.7/93.8 & 86.7/92.3 & 97.7$\pm$0.21/91.9$\pm$0.26\\
 Zipper & 79.7/94.8 & 87.4/86.1 & 98.8/98.3 & 96.0/96.1 & 95.8/96.8 & \textbf{100}/\textbf{99.5} & 93.8/95.7 & 92.2/95.7 & \textbf{100}$\pm$0.04/99.0$\pm$0.09\\
\midrule
\midrule
 \textbf{Mean} & 84.2/89.5 & 81.9/84.9 & 88.1/87.2 & 94.5/95.6 & 96.5/96.8 & 97.2/\textbf{98.3} & 91.8/96.0 & 89.0/94.0 & \textbf{98.4}$\pm$0.08/97.9$\pm$0.05\\
\bottomrule
\end{tabular}}
\end{table*}

\textbf{Quantitative Results.} The detailed results on MVTecAD are shown in Tab. \ref{tab:detailed_MVTecAD}. We also report the results under the one-for-one setting in App. Tab. \ref{tab:detailed_MVTecAD_sup}. By comparison, we can see that the performances of all baselines and SOTA one-for-one NF-based AD methods drop dramatically under the unified case. However, our HGAD outperforms all baselines under the unified case significantly. Compared with the one-for-one NF-based AD counterparts, we improve the unified AD performance from 91.8\% to 98.4\% and from 96.0\% to 97.9\%. We further indicate that our model has the same network architecture as CFLOW, and we only introduce multiple inter- and intra-class centers as extra learnable parameters. This means that our novel designs are the keys to improving the unified AD ability of the
NF-based AD methods. Moreover, our HGAD also surpasses the SOTA unified AD methods, PMAD (by 3.9\% and 2.3\%) and UniAD (by 1.9\% and 1.1\%), demonstrating our superiority. Furthermore, the results on BTAD, MVTec3D-RGB, VisA, and Union dataset (see Tab. \ref{tab:BTAD_MVTEC3D}) also verify the superiority of our method, where we outperform UniAD by 0.9\%, 9.6\%, 4.3\%, and 6.6\% in anomaly detection. 


\vspace{-0.3cm}
\begin{table*}
\caption{\textbf{Anomaly detection and localization results on BTAD, MVTec3D-RGB, VisA, and Union datasets}.}
\label{tab:BTAD_MVTEC3D}
\resizebox{1.0\linewidth}{!}{
\begin{tabular}{c||ccc|ccc|ccc}
\toprule
  \textbf{Dataset} & PaDiM  & MKD  & DRAEM & PMAD  & UniAD & OmniAL & FastFlow & CFLOW & HGAD (Ours)\\
\midrule
 \textbf{BTAD} & 93.8/96.6 & 89.7/96.2 & 91.2/91.9 & 93.8/97.3 & 94.0/97.2 & -/- & 92.9/95.3 & 93.0/96.6 & \textbf{94.9}$\pm$0.08/\textbf{98.0}$\pm$0.10\\
\midrule
  \textbf{MVTec3D-RGB} & 77.4/96.3 & 73.5/95.9 & 73.9/95.5 & 75.4/95.3 & 77.5/96.6 & -/- & 67.9/90.2 & 71.6/95.7 & \textbf{87.1}$\pm$0.16/\textbf{97.7}$\pm$0.05\\
 \midrule
  \textbf{VisA} & 86.8/97.0 & 74.2/93.9 & 85.5/90.5 & -/- & 92.8/98.1 & 87.8/96.6 & 77.2/95.1 & 88.0/95.9 & \textbf{97.1}$\pm$0.09/\textbf{98.9}$\pm$0.05\\
 \midrule
 \textbf{Union} & 79.0/91.4 & 72.1/88.9 & 66.4/82.7 & -/- & 86.9/95.5 & -/- & 57.2/78.8 & 55.7/82.9 & \textbf{93.5}$\pm$0.23/\textbf{97.5}$\pm$0.14\\
\bottomrule
\end{tabular}}
\end{table*}
\vspace{-0.3cm}

\begin{figure}[ht]
    \centering
    \includegraphics[width=1.0\linewidth]{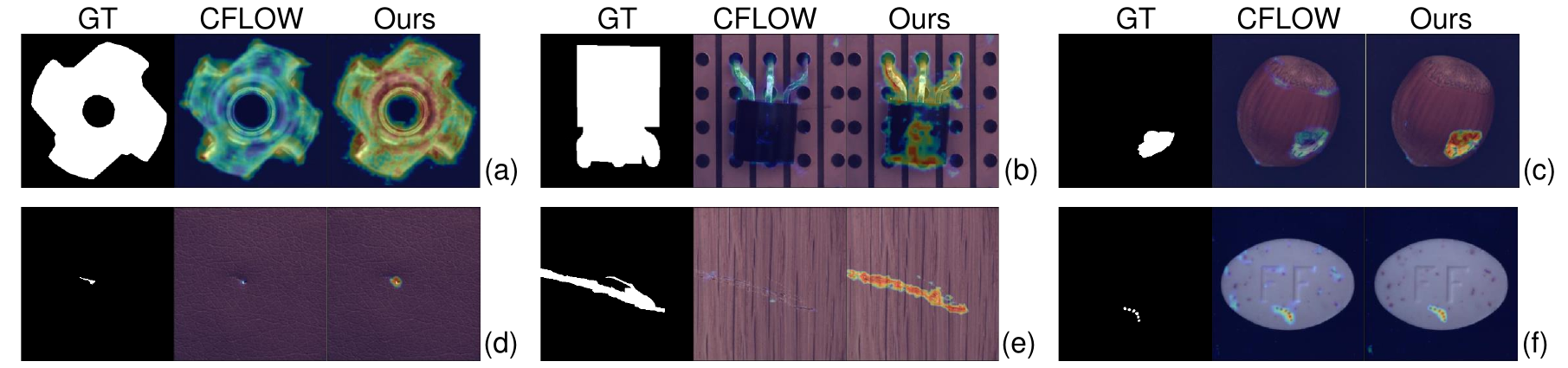}
    \caption{\textbf{Qualitative results on MVTecAD}. (a) and (b) both represent global anomalies, (c) contains large cracks, (d) shows small dints, (e) contains texture scratches, and (f) shows color anomalies.}
    \label{fig:qualitative_results}
\end{figure}

\textbf{Qualitative Results.} Fig. \ref{fig:qualitative_results} shows qualitative  results. It can be found that our approach can generate much better anomaly score maps than the one-for-one NF-based baseline CFLOW \cite{CFLOW} even for different anomaly types. More qualitative results are in the App. Fig. \ref{fig:qualitative_results_sup}.

\subsection{Ablation Studies}
\label{sec:ablation_studies}
\begin{table}[t]
\caption{\textbf{Ablation studies on MVTecAD}. \emph{SGC}, \emph{FMC}, \emph{ICG}, \emph{MIM}, and \emph{Intra} mean single center, fixed multiple centers, inter-class Gaussian mixture modeling, mutual information maximization, and intra-class mixed class centers learning, respectively.}
\vspace{-0.3cm}
\label{tab:ablation_studies}
    \begin{subtable}{0.6\linewidth}
    \centering
    \caption{Hierarchical Gaussian mixture prior.}
        \resizebox{!}{1.5cm}{\begin{tabular}{ccccc|cc}
            \toprule
            SGC & FMC & \textbf{ICG} & \textbf{MIM} & \textbf{Intra} & Det. & Loc. \\
            \midrule
            $\checkmark$ & - & - & - & - & 89.0 & 94.0\\
            - & $\checkmark$ & - & - & - & 93.1 & 95.8 \\
            - & - & $\checkmark$ & - & - & 94.5 & 96.7\\
            - & - & $\checkmark$ & $\checkmark$ & - & 96.3 & 96.6\\
            - & $\checkmark$ & - & - & $\checkmark$ & 96.3 & 97.1\\
            - & - & $\checkmark$ & $\checkmark$ & $\checkmark$ & \textbf{97.7} & \textbf{97.6} \\
            \bottomrule
        \end{tabular}}
    \end{subtable}%
    \begin{subtable}{0.4\linewidth}
    \centering
    \caption{Number of intra-class centers.}
        \resizebox{!}{1.5cm}{\begin{tabular}{c|cc}
            \toprule
            \# Centers & Det. & Loc. \\
            \midrule
            3 & 97.5 & 97.3\\
            5 & 97.6 & 97.1\\
            \textbf{10} & \textbf{97.7} & \textbf{97.6}\\
            15 & 97.6 & 97.3\\
            20 & 97.4 & 97.2\\
            \bottomrule
        \end{tabular}}
    \end{subtable}

\vspace{+0.3cm}
    \begin{subtable}{0.3\linewidth}
    \centering
    \caption{Anomaly criterion.}
        \resizebox{!}{0.9cm}{\begin{tabular}{cc|cc}
            \toprule
            \textbf{Logps} & \textbf{Entropy} & Det. & Loc. \\
            \midrule
            $\checkmark$ & - & 94.4 & 97.1\\
            - & $\checkmark$ & 96.6 & 97.0\\
            $\checkmark$ & $\checkmark$ & \textbf{97.7} & \textbf{97.6}\\
            \bottomrule
        \end{tabular}}
    \end{subtable}%
    \begin{subtable}{0.4\linewidth}
    \centering
    \caption{Hyperparameters.}
        \resizebox{!}{0.9cm}{\begin{tabular}{cc|cc||cc|cc}
            \toprule
            $\lambda_1$ & $\lambda_2$ & Det. & Loc. &  $\lambda_1$ & $\lambda_2$ & Det. & Loc.\\
            \midrule
            1 & 1 & 96.5 & 96.7 & 0.5 & 100 & 98.4 & 97.8\\
            1 & 5 & 97.6 & 97.3 & \textbf{1} & \textbf{100} & \textbf{98.4} & \textbf{97.9}\\
            1 & 10 & 97.7 & 97.6 & 5 & 100 & 98.3 & 97.9\\
            1 & 50 & 98.2 & 97.8 & 10 & 100 & 98.3 & 97.8\\
            \textbf{1} & \textbf{100} & \textbf{98.4} & \textbf{97.9} & 20 & 100 & 97.8 & 97.6\\
            \bottomrule
        \end{tabular}}
    \end{subtable}%
    \begin{subtable}{0.3\linewidth}
    \centering
    \caption{Optimization strategy.}
        \resizebox{!}{0.9cm}{\begin{tabular}{c|cc}
            \toprule
             & Det. & Loc. \\
            \midrule
             - & 96.5 & 97.0\\
            $\checkmark$ & \textbf{97.7} & \textbf{97.6}\\
            \bottomrule
        \end{tabular}}
    \end{subtable}
\end{table}

\textbf{Hierarchical Gaussian Mixture Modeling.} 1) Tab. \ref{tab:ablation_studies}a verifies our confirmation that mapping to multiple class centers is effective. With the fixed multiple centers (FMC, which is only based on fixed multiple centers but without Gaussian mixture modeling), image-level and pixel-level AUROCs can be improved by 4.1\% and 1.8\%, respectively. By employing the inter-class Gaussian mixture modeling to learn the latent multi-class distribution, the AUROCs can further be improved by 1.4\% and 0.9\%. 2) The effectiveness of Mutual information maximization (MIM) is proven in Tab. \ref{tab:ablation_studies}a, where adding MIM brings promotion by 1.8\% for detection. This shows that to better learn the complex multi-class distribution, it is necessary to endow the model class discrimination ability to avoid multiple centers collapsing into the same center. 3) Tab. \ref{tab:ablation_studies}a confirms the efficacy of intra-class mixed class centers learning. With the FMC as the baseline, introducing to learn intra-class mixed class centers could bring an increase of 1.6\% for detection and 1.0\% for localization, respectively. Finally, combining these, we form the hierarchical Gaussian mixture modeling method to achieve the best results.

\textbf{Number of Intra-Class Centers.} We conduct experiments to investigate the influence of intra-class centers in each class. The results are shown in Tab. \ref{tab:ablation_studies}b. The best performance is achieved with a moderate number: 10 class centers. A larger class center number like 20 does not bring further promotion, which may be because the class centers are saturated and more class centers are harder to train. For other datasets, we also use 10 as the number of intra-class centers.

\textbf{Anomaly Criterion.} Only taking the log-likelihood and the entropy as the anomaly criterion can achieve a good performance, while our associated criterion outperforms each criterion consistently. This illustrates that the associated anomaly scoring strategy is more conducive to guarantee that anomalies are recognized as out-of-distribution.

 \textbf{Hyperparameters.} We ablate the hyperparameters $\lambda_1$ and $\lambda_2$ in Tab. \ref{tab:ablation_studies}d. Note that the experiments in Tab. \ref{tab:ablation_studies}a, b, c, and e are conducted with $\lambda_1$ and $\lambda_2$ set to $1$ and $10$. The results in Tab. \ref{tab:ablation_studies}d show that the larger $\lambda_2$ can achieve better unified AD performance. The larger $\lambda_2$ can urge the network more focused on separating different class features into their corresponding class centers, indicating that the class discrimination ability is of vital significance to accomplish unified anomaly detection.

 \textbf{Optimization Strategy.} We found that simultaneously optimizing inter-class and intra-class centers by the beginning would bring instability. Thus, we employ a two-stage optimization strategy (see App. \ref{sec:implementation_sup} for details). The strategy can better decouple the inter-class and intra-class learning processes, thereby bringing better results.
 

\section{Conclusion}

In this paper, we focus on how to unify anomaly detection regarding multiple classes. For such a challenging task, popular normalizing flow based AD methods may fall into a ``homogeneous mapping'' issue. To address this, we propose a novel HGAD against learning the bias with three key improvements: inter-class Gaussian mixture modeling, mutual information maximization, and intra-class mixed class centers learning strategy. Under the unified AD setting, our method can improve NF-based AD methods by a large margin, and also surpass the SOTA unified AD methods. 

\section*{Acknowledgements}
 This work was supported in part by the National Natural Science Fund of China (62371295), the Shanghai Municipal Science and Technology Major Project (2021SHZDZX0102), and the Science and Technology Commission of Shanghai Municipality (22DZ2229005).

%
%
\bibliographystyle{splncs04}
\bibliography{main}

\clearpage
\appendix
\section*{Appendix}

\section{More Discussions}
\label{sec:discussion_sup}

\subsection{Necessities and Motivations}
\label{sec:motivation}
We further explain the necessities and motivations of the individual components in our method. Through the explanations in this subsection, readers can have a clearer understanding of how the overall training objective is designed and how the individual losses are motivated and connected together. 

\textbf{Inter-class Gaussian mixture modeling.} The initial motivation of our method is that NF-based AD methods usually fall into the ``homogeneous mapping'' issue when applied to the unified AD task. To address this issue, we first empirically confirm that mapping to multi-modal latent distribution is effective to prevent the model from learning the bias. The most natural method is to model multiple Gaussian distributions in the latent space. However, the fixed multiple Gaussian distribution centers still result in a relatively fixed whole distribution in the latent space, lacking adaptability. Thus, the inter-class Gaussian mixture modeling is proposed to increase the adaptability of latent distribution for better fitting the complex multi-class normal distribution. 

\textbf{Mutual information maximization.} The loss function for the inter-class Gaussian mixture modeling is in Eq. \ref{eq:loss_x}, where the logsumexp operator will sum the exp values of all classes, this means that the Eq. \ref{eq:loss_x} only has the drawing characteristic to ensure the latent features are drawn together to the whole distribution. As the class centers are randomly initialized, this may cause different class centers to collapse into the same center. Therefore, We need to further introduce class repulsion property. Then, from the mutual information perspective, we propose the mutual information maximization loss for increasing class separating ability. Furthermore, we find that using Entropy as anomaly measurement is beneficial to achieve better results (see Tab. \ref{tab:ablation_studies}c). And minimizing the inter-class entropy can also introduce the class repulsion property. Moreover, directly using entropy as an optimization item is also beneficial for the effect of entropy-based measurement. Thus, we introduce entropy loss in Eq. \ref{eq:loss_e} as a regularizer item. 

\textbf{Learning intra-class mixed class centers.} Finally, we consider that in real-world scenarios, even one object class may contain diverse normal patterns. We think modeling intra-class distribution by mixture Gaussian prior should also be beneficial for the results (see Tab. \ref{tab:ablation_studies}a). Moreover, as we further explain in Sec. \ref{sec:guaratee_way}, another consideration is to guarantee the effectiveness of anomaly determination. The inter-class Gaussian mixture modeling can't effectively guarantee the anomalies that fall into the inter-class Gaussian mixture distribution to be correctly recognized. To this end, we further model the intra-class Gaussian mixture distribution for each class to ensure that the normal distribution of each class still remains compact. Therefore, even if anomalies fall into the inter-class Gaussian mixture distribution, they are usually in the low-density regions among the inter-class class centers. So, we can still ensure that anomalies are out-of-distribution through intra-class log-likelihoods.

Therefore, although our method has four parts, each part is not arbitrarily introduced, but rather well motivated to achieve better unified AD performance. These individual losses are logically well-connected together. Our method mainly introduces a new learning objective for the NF-based AD methods, which usually doesn't increase implementation and model complexity. Thus, We think our method should be general and can be applied to various NF-based AD methods to assist them in improving the unified anomaly detection capability.

\subsection{Supervision Information}
\label{sec:sup_info}
We summarize the training samples and the supervision information required by our method and other methods in Tab. \ref{tab:sup_info}. 

\begin{table}[ht]
    \centering
    \caption{Training samples and supervision information summarization.}
    \resizebox{1.0\linewidth}{!}{\begin{tabular}{c|c|c|c|c|c|c|c|c}
    \toprule
         PaDiM & MKD & DRAEM & PMAD & UniAD & OmniAL & FastFlow & CFLOW & HGAD (Ours) \\
         \hline
         N & N & N+P & N & N & N+P & N & N & N \\
         \hline
         S & S & S & S & w/o S & w/o S & S & S & S \\
         \bottomrule
    \end{tabular}}
    \label{tab:sup_info}
\end{table}
where N means only using normal samples during training, P means also using pseudo (or synthetic) anomalies during training, S means requiring separating different classes and w/o S means not separating different classes.

If we think that using synthetic anomalies introduces anomalous information during training, DRAEM and OmniAL can also be called as supervised or self-supervised, while others are unsupervised. In addition, both UniAD and OmniAL use additional information to simulate anomalies. UniAD adds noise while OmniAL uses synthetic anomalies to learn how to reconstruct anomalies into normal during training. But our method is entirely based on learning normal feature distribution without any additional information (If synthetic anomalies can be used, our method can easily be combined with BGAD \cite{BGAD} to achieve better unified anomaly detection results). However, the methods based on synthetic anomalies may perform much worse when synthetic anomalies cannot simulate real anomalies well. This will result in limited application scenarios for such methods. For example, on the more challenging VisA dataset, our method significantly outperforms OmniAL (97.1/98.9 vs. 87.8/96.6). Compared to UniAD, results on multiple datasets, such as MVTec3D-RGB, VisA, and Union datasets, also show significant improvements (87.1 vs. 77.5, 97.1 vs. 92.8, 93.5 vs. 86.9).

Our method can be easily extended to completely unsupervised, as industrial images often have significant differences between different classes. For instance, after extracting global features, we can use a simple unsupervised clustering algorithm to divide each image into a specific class. Or we can only require few-shot samples for each class as a reference, and then compute the feature distances between each input sample to these reference samples. In this way, we can also conveniently divide each sample into the most relevant class. 

\subsection{More Discussions with ``identical shortcut''}
\label{sec:discussion_with_identical_shortcut}

 The ``identical shortcut'' is essentially caused by the leakage of abnormal information. The process of reconstruction is to remove abnormal information in the input, resulting in the failure of reconstruction in abnormal regions. But if the abnormal features are leaked into the output, this will result in the reconstruction network directly returning a copy of the input as output. This issue usually can be addressed by masking, such as the neighbor masking mechanism in UniAD \cite{UniAD}. However, the ``homogeneous mapping'' is a specific issue in normalizing flow (NF) based AD methods. In previous NF-based AD methods, the latent feature space is uni-modal. When used for unified anomaly detection, we need to map different class features to the single latent center, this may cause the model more prone to take a bias to map different input features to similar latent features. Thus, with the bias, the log-likelihoods of abnormal features will become closer to the log-likelihoods of normal features, causing normal misdetection or abnormal missing detection. We call this phenomenon as the ``homogeneous mapping'' issue, rather than casually introducing it. Moreover, as analyzed in Sec. \ref{sec:revisiting}, we provide a reasonable explanation from the perspective of the formula in normalizing flow. To address this issue, we propose the hierarchical Gaussian Mixture modeling approach, the key designs in our method are completely different from those in UniAD. As the causes and solutions of the two issues are significantly different, ``homogeneous mapping'' is not intrinsically equal to ``identical shortcut''. 

\subsection{The Way to Guarantee Anomalies Out-of-Distribution.}
\label{sec:guaratee_way}
Here, we further explain how to guarantee that anomalies are out-of-distribution. In our method, increasing inter-class distances is to ensure that the latent space has sufficient capacity to accommodate the features of multiple classes. In addition, we also model the intra-class Gaussian mixture distribution for each class to ensure that the normal distribution of each class still remains compact. Therefore, even if anomalies fall into the inter-class Gaussian mixture distribution, they are usually in the low-density regions among the inter-class class centers. So, we can still ensure that anomalies are out-of-distribution through intra-class Gaussian mixture distributions. As described in Anomaly Scoring section (sec. \ref{sec:anomaly_scoring}), we can guarantee that anomalies are recognized as out-of-distribution by combining intra-class log-likelihood and inter-class entropy to measure anomalies. Because only if the anomaly is out-of-distribution, the anomaly score based on the association of log-likelihood and entropy will be high, and the detection metrics can be better. The visualization results (decision-level results based on log-likelihood) in Fig. \ref{fig:analysis} and \ref{fig:logp_results_sup} also intuitively show that our method has fewer normal-abnormal overlaps and the normal boundary is more compact.

\subsection{Limitations} 

In this paper, we propose a novel HGAD to accomplish the unified anomaly detection task. Even if our method manifests good unified AD performance, there are still some limitations of our work. Here, we discuss two main limitations as follows:

One limitation is that our method mainly targets NF-based AD methods to improve their unified AD abilities. To this end, our method cannot be directly utilized to the other types of anomaly detection methods, such as reconstruction-based, OCC-based, embedding-based, and distillation-based approaches (see Related Work, Sec. \ref{sec:related_work}). However, we believe that the other types of anomaly detection methods can also be improved into unified AD methods, but we need to find and solve the corresponding issues in the improvement processes, such as the ``identical shortcut'' issue \cite{UniAD} in reconstruction-based AD methods. How to upgrade the other types of anomaly detection methods to unified AD methods and how to find a general approach for unified anomaly detection modeling will be the future works.


In this work, our method is mainly aimed at solving unified anomaly detection, it doesn't have the ability to directly generalize to unseen classes. Because, in our method, the new class features usually do not match the learned known multi-class feature distribution, which can lead to normal samples being misrecognized as anomalies. Generalization to unseen classes can be defined as class-agnostic anomaly detection \cite{PMAD}, where the model is trained with normal instances from multiple known classes with the objective to detect anomalies from unseen classes. In the practical industrial scenarios, models with class-agnostic anomaly detection capabilities are very valuable and necessary, because new products will continuously appear and it's cost-ineffective and inconvenient to retrain models for new products. We think our method should achieve better performance on unseen classes than previous NF-based methods due to the ability to learn more complex multi-class distribution, but it's far from solving the problem. How to design a general approach for class-agnostic anomaly detection
modeling will be the future works.

\subsection{Model Complexity}

 With the image size fixed as $256\times256$, we compare the FLOPs and learnable parameters with all competitors. In Tab. \ref{tab:complexity_comparison_sup}, we can conclude that the advantage of HGAD does not come from a larger model capacity. Compared to UniAD, our method requires fewer epochs (100 vs. 1000) and has a shorter training time.

\begin{table*}
\caption{\textbf{Complexity comparison} between our HGAD and other baseline methods.}
\label{tab:complexity_comparison_sup}
\resizebox{1.0\linewidth}{!}{
\begin{tabular}{c||ccc|cc|ccc}
\toprule
   & PaDiM  & MKD  & DRAEM & PMAD  & UniAD & FastFlow & CFLOW & HGAD (Ours)\\
\midrule
 FLOPs & 14.9G & 24.1G & 198.7G & 52G & 9.7G & 36.2G & 30.7G & 32.8G\\
 Learnable Parameters & / & 24.9M & 97.4M & 163.4M & 9.4M & 69.8M & 24.7M & 30.8M\\
 Inference Speed & 12.8fps & 23fps & 22fps & 10.8fps & 29fps & 42.7fps & 24.6fps & 24.3fps\\
 Training Epochs& / & 50 & 700 & 300 & 1000 & 400 & 200 & 100\\
\bottomrule
\end{tabular}}
\end{table*}

\subsection{Real-world Applications}

In industrial inspection scenarios, the class actually means a type of product on the production line. Unified anomaly detection can be applied to train one model to detect defects in all products, without the need to train one model for each type of product. This can greatly reduce the resource costs of training and deploying. In video surveillance scenarios, we can use one model to simultaneously detect anomalies in multiple camera scenes.

\section{Social Impacts and Ethics}
\label{sec:impact_sup}

As a unified model for unified anomaly detection, the proposed method does not suffer from particular ethical concerns or negative social impacts. All datasets used are public. All qualitative visualizations are based on industrial product images, which doesn't infringe personal privacy.

\section{Implementation Details}
\label{sec:implementation_sup}

\textbf{Optimization Strategy.} In the initial a few epochs, we only optimize with $\mathcal{L}_g$ and $\mathcal{L}_{mi}$ to form distinguishable inter-class main class centers. And then we simultaneously optimize the intra-class delta vectors and the main class centers with the overall loss $\mathcal{L}$ in Eq. \ref{eq:loss_overall}. In this way, we can better decouple the inter-class and intra-class learning processes. This strategy can make the intra-class learning become much easier, as optimizing after forming distinguishable inter-class main centers will not have the problem that many centers initially overlap with each other.

\textbf{Model Architecture.} The normalizing flow model in our method is mainly based on Real-NVP \cite{realNVP} architecture, but the convolutional subnetwork in Real-NVP is replaced with a two-layer MLP network. As in Real-NVP, the normalizing flow in our model is composed of the so-called coupling layers. All coupling layers have the same architecture, and each coupling layer is designed to tractably achieve the forward or reverse affine coupling transformation \cite{realNVP} (see Eq. \ref{eq:coupling_transformation}). Then each coupling layer is followed by a random and fixed soft permutation of channels \cite{SoftPerm} and a fixed scaling by a constant, similar to ActNorm layers introduced by \cite{Glow}. For the coupling coefficients (\emph{i.e.}, ${\rm exp}(s(x_1))$ and $t(x_1)$ in Eq. \ref{eq:coupling_transformation}), each subnetwork predicts multiplicative and additive components simultaneously, as done by \cite{realNVP}. Furthermore, we adopt the soft clamping of multiplication coefficients used by \cite{realNVP}. The layer numbers of the normalizing flow models are all 12. We add positional embeddings to each coupling layer, which are concatenated with the first half of the input features (\emph{i.e.}, $x_1$ in Eq. \ref{eq:coupling_transformation}). Then, the concatenated embeddings are sent into the subnetwork for predicting couping coefficients. The dimension of all positional embeddings is set to 256. The implementation of the normalizing flows in our model is based on the FrEIA library \url{https://github.com/VLLHD/FrEIA}.

\section{Datasets}
\label{sec:dataset_sup}

\textbf{MVTecAD.} The MVTecAD \cite{MVTec} dataset is widely used as a standard benchmark for evaluating unsupervised image anomaly detection methods. This dataset contains 5354 high-resolution images (3629 images for training and 1725 images for testing) of 15 different product categories. 5 classes consist of textures and the other 10 classes contain objects. A total of 73 different defect types are presented and almost 1900 defective regions are manually annotated in this dataset.

\textbf{BTAD.} The BeanTech Anomaly Detection dataset \cite{BTAD} is an another popular benchmark, which contains 2830 real-world images of 3 industrial products. Product 1, 2, and 3 of this dataset contain 400, 1000, and 399 training images respectively.

\textbf{MVTecAD-3D.} The MVTecAD-3D \cite{MVTec3D} dataset is recently proposed for 3D anomaly detection, which contains 4147 high-resolution 3D point cloud scans paired with 2D RGB images from 10 real-world categories. In this dataset, most anomalies can also be detected only through RGB images. Since we focus on image anomaly detection, we only use RGB images of the MVTecAD-3D dataset. We refer to this subset as MVTec3D-RGB.

\textbf{VisA.} The Visual Anomaly dataset \cite{VisA} is a recently proposed larger anomaly detection dataset compared to MVTecAD \cite{MVTec}. This dataset contains 10821 images with 9621 normal and 1200 anomalous samples. In addition to images that only contain single instance, the VisA dataset also have images that contain multiple instances. Moreover, some product categories of the VisA dataset, such as Cashew, Chewing gum, Fryum and Pipe fryum, have objects that are roughly aligned. These characteristics make the VisA dataset more challenging than the MVTecAD dataset, whose images only have single instance and are better aligned.

\section{Detailed Loss Function Derivation}
\label{sec:app_loss_derivation}
In this section, we provide the detailed derivation of the loss functions proposed in the main text, including $\mathcal{L}_g$ (Eq. \ref{eq:loss_x}), $\mathcal{L}_{mi}$ (Eq. \ref{eq:loss_y2}), and $\mathcal{L}_{in}$ (Eq. \ref{eq:loss_x_intra}).

\textbf{Derivation of $\mathcal{L}_g$.}
We use a Gaussian mixture model with class-dependent means $\mu_y$ and unit covariance $\mathbb{I}$ as the inter-class Gaussian mixture prior, which is defined as follows:
\begin{equation}
    p_Z(z) = \sum\nolimits_yp(y)\mathcal{N}(z;\mu_y,\mathbb{I})
\end{equation}
Below, we use $c_y$ as a shorthand of ${\rm log}p(y)$. Then, we can calculate the log-likelihood as follows:
\begin{align}
\label{eq:log_gmm_prior}
    {\rm log}p_Z(z) &= {\rm log}\big[\sum\nolimits_yp(y)\mathcal{N}(z;\mu_y,\mathbb{I})\big] \nonumber \\
    &= {\rm log}\big[\sum\nolimits_yp(y)(2\pi)^{-\frac{d}{2}}{\rm e}^{-\frac{1}{2}(z-\mu_y)^T(z-\mu_y)}\big] \nonumber \\
    &= -\frac{d}{2}{\rm log}(2\pi) + {\rm log}\big(\sum\nolimits_y{\rm e}^{c_y}\cdot{\rm e}^{-\frac{||z-\mu_y||^2_2}{2}}\big) \nonumber \\
    &= -\frac{d}{2}{\rm log}(2\pi) + {\rm log}\big(\sum\nolimits_y{\rm e}^{-\frac{||z-\mu_y||^2_2}{2}+c_y}\big) \nonumber \\
    &= -\frac{d}{2}{\rm log}(2\pi) + \mathop{{\rm logsumexp}}\limits_y\bigg(-\frac{||z-\mu_y||^2_2}{2}+c_y\bigg) 
\end{align}

Then, we bring the ${\rm log}p_Z(z)$ into Eq. \ref{eq:formula1} to obtain the log-likelihood ${\rm log}p_\theta(x)$ as:
\begin{equation}
    {\rm log}p_\theta(x) = -\frac{d}{2}{\rm log}(2\pi) + \mathop{{\rm logsumexp}}\limits_y\bigg(-\frac{||\varphi_\theta(x)-\mu_y||^2_2}{2}+c_y\bigg) + {\rm log}|{\rm det}J|
\end{equation}

Further, the maximum likelihood loss in Eq. \ref{eq:formula2} can be written as:
\begin{align}
    \mathcal{L}_m &= \mathbb{E}_{x \sim p(X)}[-{\rm log}p_\theta(x)] \nonumber \\
    &= \mathbb{E}_{x \sim p(X)}\bigg[-\mathop{{\rm logsumexp}}\limits_y\bigg(-\frac{||\varphi_\theta(x)-\mu_y||^2_2}{2}+c_y\bigg) - {\rm log}|{\rm det}J| + \frac{d}{2}{\rm log}(2\pi)\bigg]
\end{align}

The loss function $\mathcal{L}_g$ is actually defined as the above maximum likelihood loss $\mathcal{L}_m$ with inter-class Gaussian mixture prior.

\textbf{Extending $\mathcal{L}_g$ for Learning Intra-Class Mixed Class Centers.} When we extend the Gaussian prior $p(Z|y) = \mathcal{N}(\mu_y, \mathbb{I})$ to mixture Gaussian prior $p(Z|y) = \sum_{i=1}^Mp_i(y)\mathcal{N}(\mu_i^y, \mathbb{I})$, where $M$ is the number of intra-class latent centers, the likelihood of latent feature $z$ can be calculated as follows:
\begin{equation}
    p_Z(z) = \sum\nolimits_yp(y)\Big(\sum\nolimits_{i=1}^Mp_i(y)\mathcal{N}(\mu_i^y, \mathbb{I})\Big)
\end{equation}

Then, following the derivation in Eq. \ref{eq:log_gmm_prior}, we have:
\begin{equation}
    {\rm log}p_Z(z) = {\rm log}\Big(\sum\nolimits_yp(y)\mathop{{\rm sumexp}}\limits_i\Big[\frac{-||z-\mu_i^y||_2^2}{2} + c_i^y - \frac{d}{2}{\rm log}(2\pi)\Big]\Big)
\end{equation}
where $c_i^y$ is the shorthand of ${\rm log}p_i(y)$. The $\mathcal{L}_g$ for learning intra-class mixed class centers can be defined as:
\begin{equation}
\label{eq:loss_gmm_intra}
    \mathcal{L}_g = \mathbb{E}_{x \sim p(X)}\bigg[-{\rm log}\Big(\sum\nolimits_yp(y)\mathop{{\rm sumexp}}\limits_i\Big[\frac{-||\varphi_\theta(x)-\mu_i^y||_2^2}{2} + c_i^y - \frac{d}{2}{\rm log}(2\pi)\Big]\Big) - {\rm log}|{\rm det}J|\bigg]
\end{equation}

However, as the initial latent features $Z$ usually have large distances with the intra-class centers $\{\mu_i^y\}_{i=1}^M$, this will cause the value after ${\rm sumexp}$ operation close to $0$. After calculating the logarithm function, it's easy to cause the loss to be numerically ill-defined (NaN). Besides, we find that directly employing Eq. \ref{eq:loss_gmm_intra} for learning intra-class mixed class centers will lead to much worse results, as we need to simultaneously optimize all intra-class centers of all classes to fit the inter-class Gaussian mixture prior. To this end, we propose to decouple the inter-class Gaussian mixture prior fitting and the intra-class latent centers learning. The loss function of learning intra-class mixed class centers is defined in Eq. \ref{eq:loss_intra_sup}.

\textbf{Derivation of $\mathcal{L}_{mi}$.} We first derive the general format of the mutual information loss in Eq. \ref{eq:loss_y1} as follows:
\begin{align}
\label{eq:loss_mi_derivation}
    \mathcal{L}_{mi} &= -I(Y,Z) = -H(Y) + H(Y|Z) = -H(Y) - H(Z) + H(Y,Z) \nonumber \\
    &= -H(Y) - \mathbb{E}_{x \sim p(X)}\Big[-{\rm log}\big(\sum\nolimits_yp(y)p(\varphi_\theta(x)|y)\big)\Big] \nonumber \\
    &+ \mathbb{E}_{(x,y) \sim p(X,Y)}[-{\rm log}(p(y)p(\varphi_\theta(x)|y))] \nonumber \\
    &= -H(Y) - \mathbb{E}_{(x,y) \sim p(X,Y)}\bigg[{\rm log}\frac{p(y)p(\varphi_\theta(x)|y)}{\sum\nolimits_{y^\prime}p(y^\prime)p(\varphi_\theta(x)|y^\prime)}\bigg] \nonumber \\
    &= -\mathbb{E}_{y \sim p(Y)}[-{\rm log}p(y)] - \mathbb{E}_{(x,y) \sim p(X,Y)}\bigg[{\rm log}\frac{p(y)p(\varphi_\theta(x)|y)}{\sum\nolimits_{y^\prime}p(y^\prime)p(\varphi_\theta(x)|y^\prime)}\bigg]
\end{align}

 Then, by replacing $p(\varphi_\theta(x)|y)$ with $\mathcal{N}(\varphi_\theta(x);\mu_y, \mathbb{I})$ in the mutual information loss, we can derive the following practical loss format for the second part of Eq. \ref{eq:loss_mi_derivation}. We also use $c_y$ as a shorthand of ${\rm log}p(y)$.
\begin{align}
\label{eq:loss_mi_second_part}
    &-\mathbb{E}_{(x,y) \sim p(X,Y)}\bigg[{\rm log}\frac{p(y)p(\varphi_\theta(x)|y)}{\sum\nolimits_{y^\prime}p(y^\prime)p(\varphi_\theta(x)|y^\prime)}\bigg] \nonumber \\
    &= -\mathbb{E}_{(x,y) \sim p(X,Y)}\bigg[{\rm log}\frac{p(y)\mathcal{N}(\varphi_\theta(x);\mu_y, \mathbb{I})}{\sum\nolimits_{y^\prime}p(y^\prime)\mathcal{N}(\varphi_\theta(x);\mu_{y^\prime}, \mathbb{I})}\bigg] \nonumber \\
     &= -\mathbb{E}_{(x,y) \sim p(X,Y)}\bigg[{\rm log}\frac{(2\pi)^{-\frac{d}{2}}{\rm e}^{-\frac{1}{2}(\varphi_\theta(x)-\mu_y)^T(\varphi_\theta(x)-\mu_y)} \cdot {\rm e}^{c_y}}{\sum\nolimits_{y^\prime}(2\pi)^{-\frac{d}{2}}{\rm e}^{-\frac{1}{2}(\varphi_\theta(x)-\mu_{y^\prime})^T(\varphi_\theta(x)-\mu_{y^\prime})} \cdot {\rm e}^{c_{y^\prime}}}\bigg] \nonumber \\
     &= -\mathbb{E}_{(x,y) \sim p(X,Y)}\Bigg[{\rm log}\frac{{\rm e}^{-\frac{||\varphi_\theta(x)-\mu_y||^2_2}{2} + c_y}}{\sum\nolimits_{y^\prime} {\rm e}^{-\frac{||\varphi_\theta(x)-\mu_{y^\prime}||^2_2}{2} + c_{y^\prime}} }\Bigg] \nonumber \\
     &= -\mathbb{E}_{(x,y) \sim p(X,Y)}\bigg[\mathop{{\rm logsoftmax}}\limits_y \bigg(-\frac{||\varphi_\theta(x) - \mu_{y^\prime}||^2_2}{2} + c_{y^\prime}\bigg)\bigg] 
\end{align}

By replacing Eq. \ref{eq:loss_mi_second_part} back to the Eq. \ref{eq:loss_mi_derivation}, we can obtain the following practical loss format of the mutual information loss. 
\begin{align}
    \mathcal{L}_{mi} &= -\mathbb{E}_{y \sim p(Y)}[-{\rm log}p(y)] - \mathbb{E}_{(x,y) \sim p(X,Y)}\bigg[\mathop{{\rm logsoftmax}}\limits_y \bigg(-\frac{||\varphi_\theta(x) - \mu_{y^\prime}||^2_2}{2} + c_{y^\prime}\bigg)\bigg] \nonumber \\
    &= -\mathbb{E}_{y \sim p(Y)}[-c_y] - \mathbb{E}_{(x,y) \sim p(X,Y)}\bigg[\mathop{{\rm logsoftmax}}\limits_y \bigg(-\frac{||\varphi_\theta(x) - \mu_{y^\prime}||^2_2}{2} + c_{y^\prime}\bigg)\bigg] \nonumber \\
    &= -\mathbb{E}_{(x,y) \sim p(X,Y)}\bigg[\mathop{{\rm logsoftmax}}\limits_y \bigg(-\frac{||\varphi_\theta(x) - \mu_{y^\prime}||^2_2}{2} + c_{y^\prime}\bigg) - c_y\bigg]
\end{align}

\textbf{Intra-Class Mixed Class Centers Learning Loss.} The loss function for learning the intra-class class centers is actually the same as the $\mathcal{L}_g$ in Eq. \ref{eq:loss_x}. But we note that we need to replace the class centers with the intra-class centers: $\mu_i^y = \{\mu_1^y + \Delta\mu_i^y\}_{i=1}^M$, and the ${\rm sum}$ operation is performed on all intra-class centers $\mu_i^y$ within the corresponding class $y$. Another difference is that we need to detach the main center $\mu_1^y$ from the gradient graph and only optimize the delta vectors. The loss function can be written as:
\begin{equation}
\label{eq:loss_intra_sup}
    \mathcal{L}_{in} = \mathbb{E}_{(x,y) \sim p(X,Y)}\bigg[-\mathop{{\rm logsumexp}}\limits_i\bigg(-\frac{||\varphi_\theta(x)-(SG[\mu^y_1]+\Delta\mu^y_i)||^2_2}{2}+c^y_i\bigg)-{\rm log}|{\rm det}J|\bigg]
\end{equation}

Finally, we note that the use of ${\rm logsumexp}$ and ${\rm logsoftmax}$ pytorch operations above is quite important. As the initial $||\varphi_\theta(x)-\mu_y||^2_2/2$ distance values are usually large, if we explicitly perform the ${\rm exp}$ and then ${\rm log}$ operations, the values will become too large and the loss will be numerically ill-defined (NaN).

\section{An Information-Theoretic View}
\label{sec:information_view}
Information theory \cite{Information} is an important theoretical foundation for explaining deep learning methods. The well-known \emph{Information Bottleneck principle} \cite{IB1, IB2, IB3, IB4} is also rooted from the information theory, which provides an explanation for representation learning as the trade-off between information compression and informativeness retention. Below, we denote the input variable as $X$, the latent variable as $Z$, and the class variable as $Y$. Formally, in this theory, supervised deep learning attempts to minimize the mutual information $I(X,Z)$ between the input $X$ and the latent variable $Z$ while maximizing the mutual information $I(Z,Y)$ between Z and the class $Y$:
\begin{equation}
    \mathop{{\rm min}} I(X,Z) - \alpha I(Z,Y)
\end{equation}
where the hyperparameter $\alpha > 0$ controls the trade-off between compression (\emph{i.e.}, redundant information) and retention (\emph{i.e.}, classification accuracy).

In this section, we will show that our method can be explained by the \emph{Information Bottleneck principle} with the learning objective ${\rm min} I(X, Z_{\mathcal{E}}) - \alpha I(Z, Y)$, where $Z_{\mathcal{E}} = \varphi_\theta(X + \mathcal{E})$ and $p(\mathcal{E}) = \mathcal{N}(0, \sigma^2\mathbb{I})$ is Gaussian with mean zero and covariance $\sigma^2\mathbb{I}$. First, we derive $I(X, Z_{\mathcal{E}})$ as follows:
\begin{align}
    I(X, Z_{\mathcal{E}}) &= I(Z_{\mathcal{E}}, X) = H(Z_{\mathcal{E}}) - H(Z_{\mathcal{E}}|X) \nonumber \\
    &= \underbrace{\mathbb{E}_{x \sim p(X), \epsilon \sim p(\mathcal{E})}[-{\rm log}p(\varphi_\theta(x+\epsilon))]}_{:=A} + \underbrace{\mathbb{E}_{x \sim p(X), \epsilon \sim p(\mathcal{E})}[{\rm log}p(\varphi_\theta(x+\epsilon)|x)]}_{:=B}
\end{align}

To approximate the second item ($B$), we can replace the condition $x$ with $\varphi_\theta(x)$, because $\varphi_\theta$ is bijective and both conditions convey the same information \cite{IB-INN}.
\begin{equation}
    B = \mathbb{E}_{x \sim p(X), \epsilon \sim p(\mathcal{E})}[{\rm log}p(\varphi_\theta(x+\epsilon)|x)] = \mathbb{E}_{x \sim p(X), \epsilon \sim p(\mathcal{E})}[{\rm log}p(\varphi_\theta(x+\epsilon)|\varphi_\theta(x))]
\end{equation}
We can linearize $\varphi_\theta(x+\epsilon)$ by its first order Taylor expansion: $\varphi_\theta(x + \epsilon) = \varphi_\theta(x) + J\epsilon + \mathcal{O}(\epsilon^2)$, where the matrix $J = \bigtriangledown_x\varphi_\theta(x)$ is the Jacobian matrix of the bijective transformation ($z = \varphi_\theta(x)$ and $x = \varphi_\theta^{-1}(z)$). Then, we have:
\begin{align}
    B &= \mathbb{E}_{x \sim p(X), \epsilon \sim p(\mathcal{E})}[{\rm log}p(\varphi_\theta(x) + J\epsilon + \mathcal{O}(\epsilon^2)|\varphi_\theta(x))] \nonumber \\
    & = \mathbb{E}_{x \sim p(X), \epsilon \sim p(\mathcal{E})}[{\rm log}p(\varphi_\theta(x) + J\epsilon|\varphi_\theta(x))] + \mathbb{E}_{\epsilon \sim p(\mathcal{E})}[\mathcal{O}(\epsilon^2)] \nonumber \\
    &=\mathbb{E}_{x \sim p(X), \epsilon \sim p(\mathcal{E})}[{\rm log}p(\varphi_\theta(x) + J\epsilon|\varphi_\theta(x))] + \mathcal{O}(\sigma^2)
\end{align}
where the $\mathbb{E}_{\epsilon \sim p(\mathcal{E})}[\mathcal{O}(\epsilon^2)]$ is actually the covariance of $p(\mathcal{E}) = \mathcal{N}(0, \sigma^2\mathbb{I})$, thus can be replaced with $\mathcal{O}(\sigma^2)$. Since $p(\mathcal{E})$ is Gaussian with mean zero and covariance $\sigma^2\mathbb{I}$, the conditional distribution is Gaussian with mean $\varphi_\theta(x)$ and covariance $\sigma^2JJ^T$. Then, we have:
\begin{align}
    B &= \mathbb{E}_{x \sim p(X), \epsilon \sim p(\mathcal{E})}[{\rm log}\mathcal{N}(\varphi_\theta(x) + J\epsilon; \varphi_\theta(x), \sigma^2JJ^T)] + \mathcal{O}(\sigma^2) \nonumber \\
    &= \mathbb{E}_{x \sim p(X), \epsilon \sim p(\mathcal{E})}[{\rm log}((2\pi)^{-\frac{d}{2}} \cdot (|\sigma^2JJ^T|)^{-\frac{1}{2}} \cdot {\rm e}^{-\frac{1}{2}\frac{1}{\sigma^2}\epsilon^T\epsilon})] + \mathcal{O}(\sigma^2) \nonumber \\
    &= \mathbb{E}_{x \sim p(X)}[-\frac{1}{2}{\rm log} (|\sigma^2JJ^T|)] - \frac{d}{2}{\rm log}(2\pi) -\frac{1}{2\sigma^2}\mathbb{E}_{\epsilon \sim p(\mathcal{E})}[\epsilon^T\epsilon] + \mathcal{O}(\sigma^2) \nonumber \\
    &= \mathbb{E}_{x \sim p(X)}[-\frac{1}{2}{\rm log} (|JJ^T|)] -d{\rm log}(\sigma)- \frac{d}{2}{\rm log}(2\pi) -\frac{1}{2\sigma^2}\mathcal{O}(\sigma^2) + \mathcal{O}(\sigma^2) \nonumber \\
    &= \mathbb{E}_{x \sim p(X)}[-{\rm log} |{\rm det} J|] -d{\rm log}(\sigma)- \frac{d}{2}{\rm log}(2\pi) -\frac{1}{2\sigma^2}\mathcal{O}(\sigma^2) + \mathcal{O}(\sigma^2) 
\end{align}
For the first item (A), we can use the derivation in Eq. \ref{eq:log_gmm_prior}.
\begin{align}
    A &= \mathbb{E}_{x \sim p(X), \epsilon \sim p(\mathcal{E})}[-{\rm log}p(\varphi_\theta(x+\epsilon))] \nonumber \\
    &= \mathbb{E}_{x \sim p(X), \epsilon \sim p(\mathcal{E})}\Big[\frac{d}{2}{\rm log}(2\pi)-\mathop{{\rm logsumexp}}\limits_y\Big(-\frac{||\varphi_\theta(x+\epsilon)-\mu_y||^2_2}{2}+c_y\Big)\Big] \nonumber \\
    &= \mathbb{E}_{x \sim p(X), \epsilon \sim p(\mathcal{E})}\Big[-\mathop{{\rm logsumexp}}\limits_y\Big(-\frac{||\varphi_\theta(x+\epsilon)-\mu_y||^2_2}{2}+c_y\Big)\Big] + \frac{d}{2}{\rm log}(2\pi)
\end{align}

Finally, we put the above derivations together and drop the constant items and the items that vanish with rate $\mathcal{O}(\sigma^2)$ as $\sigma \rightarrow 0$. The $I(X, Z_{\mathcal{E}})$ becomes:
\begin{equation}
    I(X, Z_{\mathcal{E}}) = \mathbb{E}_{x \sim p(X), \epsilon \sim p(\mathcal{E})}\Big[-\mathop{{\rm logsumexp}}\limits_y\Big(-\frac{||\varphi_\theta(x+\epsilon)-\mu_y||^2_2}{2}+c_y\Big) - {\rm log}|{\rm det}J|\Big]
\end{equation}

We can find that the $I(X, Z_{\mathcal{E}})$ has the same formula as the loss $\mathcal{L}_g$ except the constant item $\frac{d}{2}{\rm log}(2\pi)$, and $I(Z,Y) = I(Y,Z) = -\mathcal{L}_{mi}$ (see Eq. \ref{eq:loss_mi_derivation}). Thus, the learning objective ${\rm min}I(X, Z_{\mathcal{E}}) - \alpha I(Z,Y)$ in \emph{Information Bottleneck principle} can be converted to $\mathcal{L}_g + \alpha \mathcal{L}_{mi}$, which is the first half part of the training loss in Eq. \ref{eq:loss_overall}.

From the \emph{Information Bottleneck principle} perspective, we can explain our method: it attempts to minimize the mutual information $I(X,Z_{\mathcal{E}})$ between $X$ and $Z_{\mathcal{E}}$, forcing the model to ignore the irrelevant aspects of $X + \mathcal{E}$ which do not contribute to fit the latent distribution and only increase the potential for overfitting. Therefore, the $\mathcal{L}_g$ loss function actually endows the normalizing flow model with the compression ability for establishing correct invertible mappings between input $X$ and the latent Gaussian mixture prior $Z$, which is effective to prevent the model from learning the ``homogeneous mapping". Simultaneously, it encourages to maximize the mutual information $I(Y,Z)$ between $Y$ and $Z$, forcing the model to map different class features to their corresponding class centers which can contribute to class discriminative ability.

\section{Additional Results}
\label{sec:results_sup}

\textbf{Quantitative Results Under the One-for-one Setting.} In Tab. \ref{tab:detailed_MVTecAD_sup}, we report the detailed results of anomaly detection and localization on MVTecAD \cite{MVTec} under the one-for-one setting. We can find that all baselines achieve excellent results under the one-for-one setting, but their performances drop dramatically under the unified case (see Tab. \ref{tab:detailed_MVTecAD} in the main text). For instance, the strong baseline, DRAEM, suffers from a drop of 9.9\% and 10.1\%. The performance of the previous SOTA NF-based AD method, FastFlow, drops by 7.6\% and 2.5\%. This demonstrates that the unified anomaly detection is quite more challenging than the conventional one-for-one anomaly detection task, and current SOTA AD methods cannot be directly applied to the unified AD task well. Thus, how to improve the unified AD ability for AD methods should be further studied. On the other hand, compared with reconstruction-based AD methods (\emph{e.g}, DRAEM \cite{DRAEM}), NF-based AD methods have less performance degradation when directly applied to the unified case, indicating that NF-based approaches may be a more suitable way for the unified AD modeling than the reconstruction-based approaches.

\begin{table*}
\caption{\textbf{Anomaly detection and localization results on MVTecAD}. All methods are evaluated under the one-for-one setting. $\cdot/\cdot$ means the image-level and pixel-level AUROCs.}
\label{tab:detailed_MVTecAD_sup}
\resizebox{1.0\linewidth}{!}{
\begin{tabular}{c||ccc|cc|cc}
\toprule
\multirow{2}*{\textbf{Category}} & \multicolumn{3}{c|}{\textbf{Baseline Methods}} & \multicolumn{2}{c|}{\textbf{Unified Methods}} & \multicolumn{2}{c}{\textbf{NF Based Methods}}\\
  & PaDiM & MKD  & DRAEM & PMAD & UniAD & FastFlow & CFLOW \\
\midrule
\midrule
 Carpet & 99.8/99.0 & 79.3/95.6 & 97.0/95.5 & 99.7/98.8 & 99.9/98.0 & 100/99.4 & 100/99.3 \\
  Grid  & 96.7/97.1 & 78.0/91.8 & 99.9/99.7 & 97.7/96.3 & 98.5/94.6  & 99.7/98.3 & 97.6/99.0 \\
  Leather & 100/99.0 & 95.1/98.1 & 100/98.6 & 100/99.2 & 100/98.3  & 100/99.5 & 97.7/99.7 \\
  Tile & 98.1/94.1 & 91.6/82.8 & 99.6/99.2 & 100/94.4 & 99.0/91.8 & 100/96.3 & 98.7/98.0 \\
  Wood  & 99.2/94.1 & 94.3/84.8 & 99.1/96.4 & 98.0/93.3 & 97.9/93.4  & 100/97.0 & 99.6/96.7 \\
\midrule
  Bottle & 99.9/98.2 & 99.4/96.3 & 99.2/99.1 & 100/98.4 & 100/98.1  & 100/97.7 & 100/99.0 \\
 Cable  & 92.7/96.7 & 89.2/82.4 & 91.8/94.7 & 98.0/97.5 & 97.6/96.8  & 100/98.4 & 100/97.6 \\
 Capsule & 91.3/98.6 & 80.5/95.9 & 98.5/94.3 & 89.8/98.6 & 85.3/97.9  & 100/99.1 & 99.3/99.0 \\
 Hazelnut & 92.0/98.1 & 98.4/94.6 & 100/99.7 & 100/98.8 & 99.9/98.8  & 100/99.1 & 96.8/98.9\\
 Metal nut & 98.7/97.3 & 73.6/86.4 & 98.7/99.5 & 99.2/97.5 & 99.0/95.7  & 100/98.5 & 91.9/98.6 \\
 Pill & 93.3/95.7 & 82.7/89.6 & 98.9/97.6 & 94.3/95.5 & 88.3/95.1  & 99.4/99.2 & 99.9/99.0 \\
 Screw & 85.8/98.4 & 83.3/96.0 & 93.9/97.6 & 73.9/91.4 & 91.9/97.4  & 97.8/99.4 & 99.7/98.9 \\
 Toothbrush & 96.1/98.8 & 92.2/96.1 & 100/98.1 & 91.4/98.2 & 95.0/97.8  & 94.4/98.9 & 95.2/99.0 \\
 Transistor & 97.4/97.6 & 85.6/76.5 & 93.1/90.9 & 99.8/97.8 & 100/98.7 & 99.8/97.3 & 99.1/98.0 \\
 Zipper & 90.3/98.4 & 93.2/93.9 & 100/98.8 & 99.5/96.7 & 96.7/96.0  & 99.5/98.7 & 98.5/99.1 \\
\midrule
\midrule
 \textbf{Mean} & 95.5/97.4 & 87.8/90.7 & 98.0/97.3 & 96.1/96.8 & 96.6/96.6 & 99.4/98.5 & 98.3/98.6 \\
\bottomrule
\end{tabular}}
\end{table*}

\begin{figure*}
    \centering
    \includegraphics[width=1.0\linewidth]{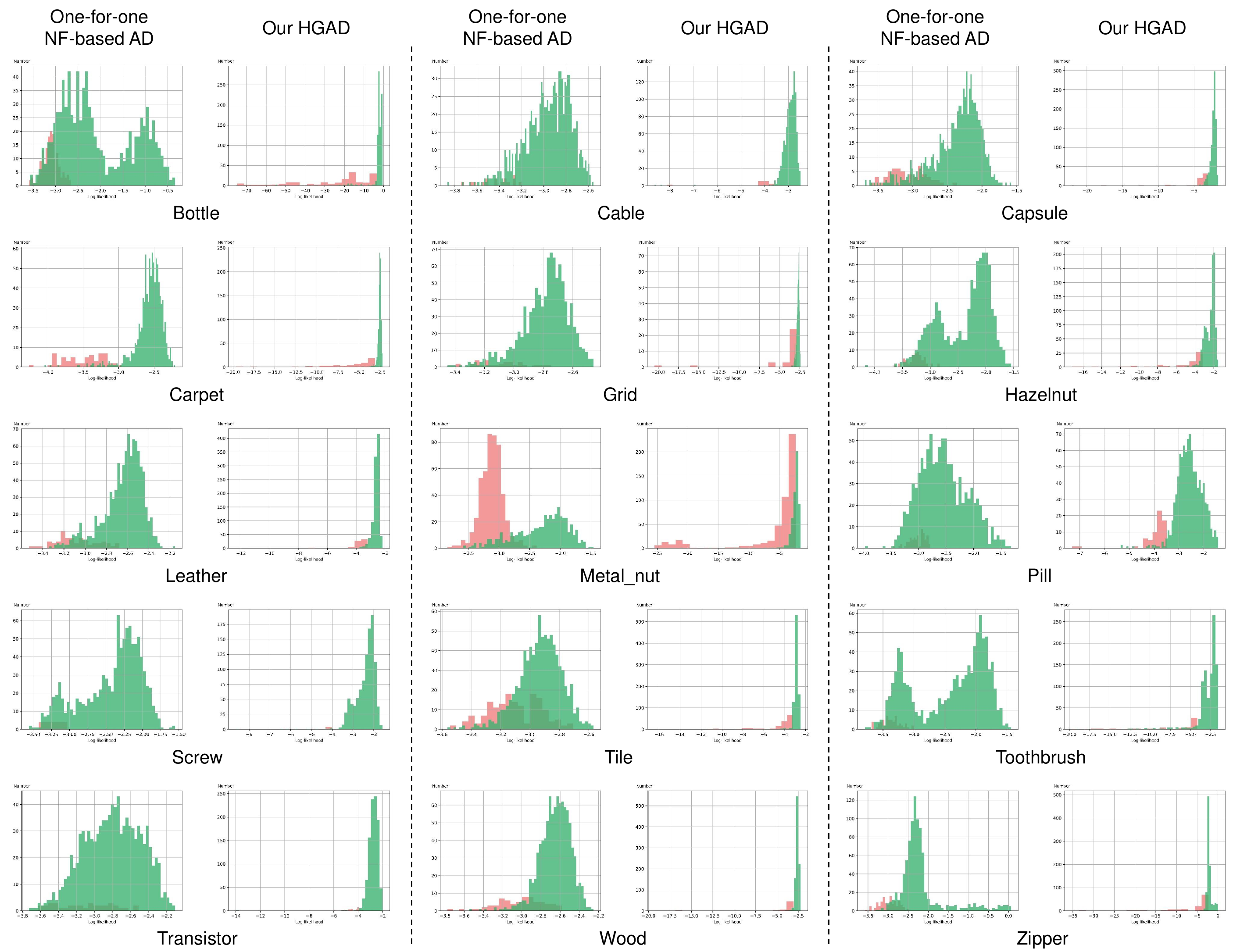}
    \caption{\textbf{Log-likelihood histograms on MVTecAD}. All categories are from the MVTecAD dataset.}
    \label{fig:logp_results_sup}
\end{figure*}

\textbf{Log-likelihood Histograms.} In Fig. \ref{fig:logp_results_sup}, we show log-likelihoods generated by the one-for-one NF-based AD method and our method. All categories are from the MVTecAD dataset. The visualization results can empirically verify our speculation that the one-for-one NF-based AD methods may fall into the ``homogeneous mapping" issue, where the normal and abnormal log-likelihoods are highly overlapped.

\begin{figure*}
    \centering
    \includegraphics[width=0.9\linewidth]{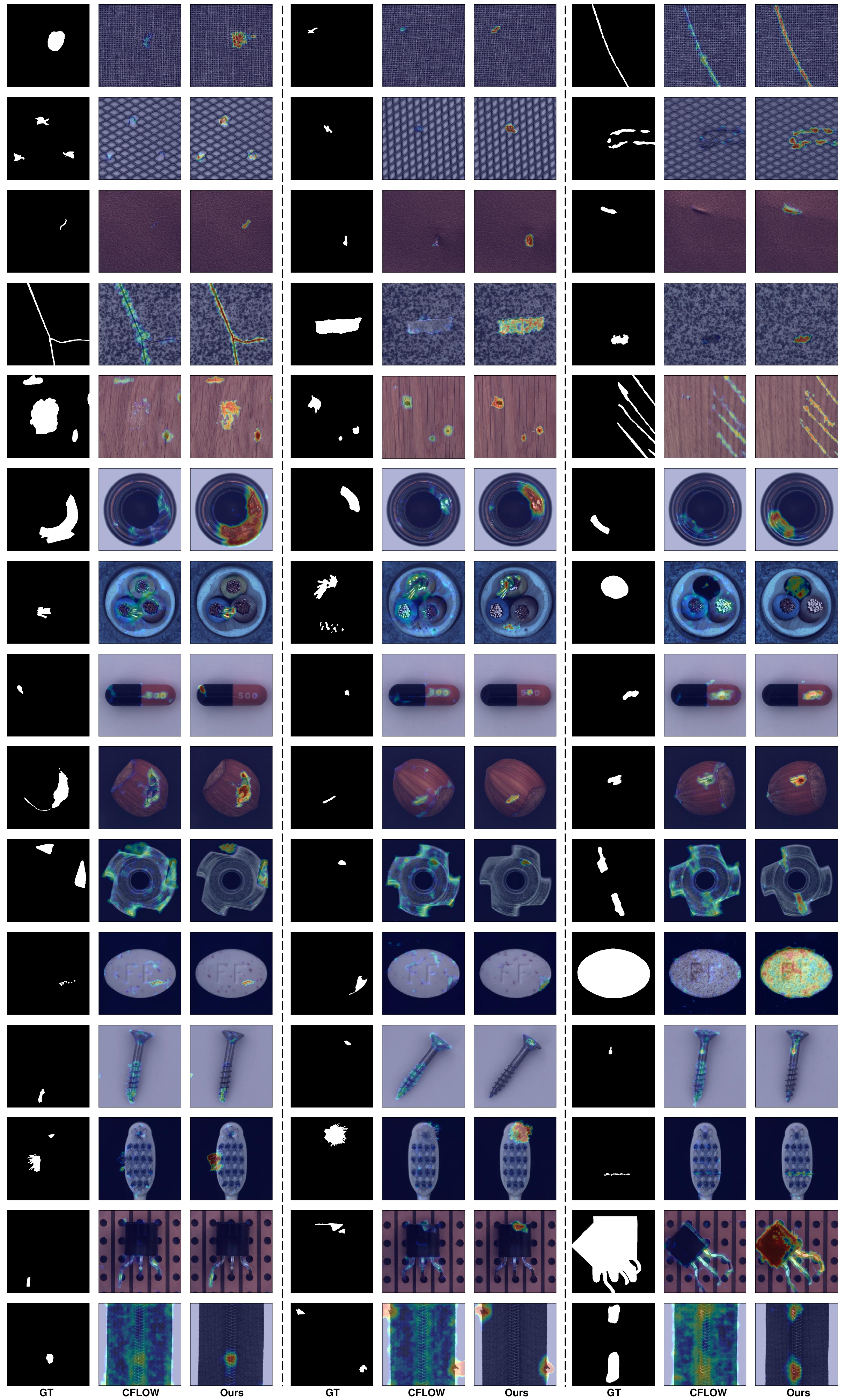}
    \caption{\textbf{Qualitative results on MVTecAD}. More visualization of anomaly localization maps generated by our method on industrial inspection data. All examples are from the MVTecAD dataset.}
    \label{fig:qualitative_results_sup}
\end{figure*}

\textbf{Qualitative Results.} We present in Fig. \ref{fig:qualitative_results_sup} additional anomaly localization results of categories with different anomalies in the MVTecAD dataset. It can be found that our approach can generate much better anomaly score maps that the one-for-one NF-based baseline CFLOW \cite{CFLOW} even for different categories from the MVTecAD dataset.

\end{document}